\documentclass{article}

    \PassOptionsToPackage{numbers, compress}{natbib}
\usepackage[preprint]{neurips_2026}

\usepackage[utf8]{inputenc} % allow utf-8 input
\usepackage[T1]{fontenc}    % use 8-bit T1 fonts
\usepackage{hyperref}       % hyperlinks
\usepackage{url}            % simple URL typesetting
\usepackage{booktabs}       % professional-quality tables
\usepackage{amsfonts}       % blackboard math symbols
\usepackage{nicefrac}       % compact symbols for 1/2, etc.
\usepackage{microtype}      % microtypography
\usepackage{xcolor}         % colors

\usepackage{enumitem}
\usepackage{graphicx}
\usepackage{wrapfig}
\usepackage{amsmath}
\usepackage{amsthm}
\usepackage{float}
\usepackage{placeins}

\graphicspath{{./}{figures/}}

\title{Elicitation Matters: How Prompts and Query Protocols Shape LLM Surrogates under Sparse Observations}

\author{%
  Ge Lei \\
  Dyson School of Design Engineering\\
  Imperial College London\\
  Exhibition Road, London SW7 2AZ \\
  \texttt{g.lei23@imperial.ac.uk} \\
  \And
  Samuel J. Cooper \\
  Dyson School of Design Engineering\\
  Imperial College London \\
  Exhibition Road, London SW7 2AZ \\
  \texttt{samuel.cooper@imperial.ac.uk} \\
}

\newcommand{\pw}{\textsc{Pointwise}}
\newcommand{\joint}{\textsc{Joint}}
\newcommand{\neutral}{\textsc{Neutral}}
\newcommand{\underdet}{\textsc{Underdetermined}}

\newcommand{\mcvar}{v_{\mathcal{H}}}

\theoremstyle{definition}
\newtheorem{definition}{Definition}[section]
\newtheorem{remark}{Remark}[section]
\theoremstyle{plain}
\newtheorem{proposition}{Proposition}[section]

\begin{document}

\maketitle

\begin{abstract}
% LLM surrogates in optimization are not specified by observations alone: they are elicited through prompt text and query protocol. This matters most in low-data regimes, where sparse observations admit many plausible objectives and the acquisition function depends on both predictions and uncertainty. We study the optimizer-facing predictive object induced by an LLM, which we call a protocol-conditioned surrogate belief. Across controlled sparse function-inference tasks, we show that prompt and query protocol are not mere interface details: they define the surrogate belief seen by the optimizer and measurably change uncertainty, consistency, belief revision, acquisition choices, and regret. Correct problem cues act as effective priors and improve the elicited surrogate, while incorrect cues systematically mislead the surrogate. Pointwise and joint elicitation also induce distinct beliefs: pointwise querying better preserves ambiguity-sensitive uncertainty and observation faithfulness, whereas joint querying yields more coherent but less constraint-faithful completions. Under sequential evidence, the surrogate revises in non-monotonic and order-sensitive ways. In BO, these differences alter acquisition behavior and regret. Thus, prompt design and query protocol are not implementation details; they are part of the surrogate specification.

Large language models are increasingly used as surrogate models for low-data optimization, but their optimizer-facing prediction and its uncertainty remain poorly understood. We study the surrogate belief elicited from an LLM under sparse observations, showing that it depends strongly on prompt text and query protocol. We introduce an uncertainty-alignment criterion that measures whether model uncertainty tracks residual ambiguity among sample-consistent functions. Across controlled inference tasks and Bayesian optimization studies, we find that structural prompts act as effective priors, POINTWISE and JOINT querying induce different beliefs, and sequential evidence leads to non-monotonic, order-sensitive confidence updates. These effects change downstream acquisition decisions and regret, showing that elicitation protocol is part of the LLM surrogate specification, not a formatting detail.
\end{abstract}

\section{Introduction}

Large language models can perform nontrivial in-context learning (ICL) without gradient updates \citep{brown2020language,garg2022can,akyurek2022learning}. This has motivated interest in using them as surrogate regressors inside experiment design and optimization workflows \citep{hao2024large,liu2024large,chang2025llinbo}. Most evaluations of such systems focus on predictive accuracy and, increasingly, calibration against realized error \citep{geng2024survey,liu2025uncertainty}. Those are useful diagnostics, but they do not fully capture the regime that matters most for early-stage optimization.

In the low-data setting, the objective is typically underdetermined: a small observation set constrains many plausible completions. In this regime, the optimizer needs more than a point estimate. It needs a notion of uncertainty that captures the remaining degrees of freedom in objectives still consistent with the observations. It should also update this uncertainty sensibly when new evidence is introduced. Classical surrogates make these commitments explicit through function classes, priors, and posterior updates \citep{williams2006gaussian,snoek2012practical}. LLM surrogates instead often inherit implicit regularities from pretraining, prompt wording, and query protocol.

For LLM surrogates, the optimizer-facing belief is elicited through language. However, linguistic competence is not the same as surrogate behavior. An LLM may correctly explain descriptors such as \emph{monotonic}, \emph{quadratic}, or \emph{underdetermined}, but that does not imply that these descriptors constrain the numerical predictions or uncertainty estimates it returns. What matters for optimization is the operational belief induced by the full elicitation procedure.

Prior work has mainly evaluated LLMs by their numerical prediction accuracy or by their performance inside Bayesian optimization loops. 
We instead ask what surrogate the optimizer actually receives from an LLM under sparse observations. 
Given the same observations $D$, prompt text $p$ and query protocol $\pi$ can change the predictions, uncertainty estimates, and induced predictive distributions seen by the optimizer. 
We call this optimizer-facing object a \emph{protocol-conditioned surrogate belief}. 
Here, \emph{belief} is operational: it denotes observable predictive behavior, not the model's internal state or an approximate Bayesian posterior. 
Our goal is not to show that prompts can change LLM outputs, but to characterize the surrogate belief induced by the full elicitation procedure: how prompt text and query protocol shape predictions, uncertainty, induced predictive distributions, downstream optimization decisions, and sequential updates under new evidence. We make four contributions:

% This distinction is important because prompt and protocol choices do not merely change surface outputs. They can change the surrogate belief that drives acquisition. Such changes can affect uncertainty alignment, cross-query consistency, belief revision under new evidence or structural constraints, acquisition behavior, and downstream regret. Thus, our focus is not prompt sensitivity in isolation, but whether elicitation choices reshape the optimizer-facing object on which decisions are based.

% This framing also separates verbal reasoning from numerical surrogate behavior. An LLM may correctly explain descriptors such as \emph{monotonic}, \emph{quadratic}, or \emph{underdetermined}, but that does not mean those descriptors guide the computation that produces its predictions and uncertainties. For optimization, the relevant question is behavioral: when a structural descriptor or a new observation is added, does the predicted curve, the uncertainty pattern, and the next query choice change in a way that reflects the added information?

\begin{itemize}[leftmargin=*]
    % \item We introduce \emph{protocol-conditioned surrogate belief}: the predictions and uncertainty estimates that an optimizer receives from an LLM under a given set of observations, prompt, and query protocol.

    \item We propose an \emph{uncertainty-alignment} criterion that compares
    model-side uncertainty with residual sample-consistent ambiguity.

    \item We show that prompt language acts as an effective prior: correct
    structural cues help, while incorrect cues systematically mislead the
    surrogate.

    \item We show that POINTWISE and JOINT querying induce distinct surrogate
beliefs: POINTWISE is more ambiguity-sensitive and observation-faithful, while
JOINT is more coherent but less constraint-faithful.

  \item We show that under sequential evidence, LLM surrogate beliefs exhibit a Dunning–Kruger-like dip-and-recovery in confidence, with evidence order shifting the confidence valley.
\end{itemize}

\section{Problem Setting}
\label{sec:setup}

We study sparse-observation function inference: an LLM observes a small set of $(x,y)$ pairs from an unknown objective and predicts values at unseen points. Our interest is the predictive object exposed to a downstream optimizer, and whether its uncertainty tracks the ambiguity left by sparse data.

\subsection{Sparse-observation function inference}
Let $f^{\star}:\mathcal{X}\rightarrow\mathbb{R}$ denote the unknown objective. After $t$ evaluations, the observation set is
\begin{equation}
D_t = \{(x_i,y_i)\}_{i=1}^{t}, \qquad y_i=f^{\star}(x_i).
\end{equation}
For a reference class $\mathcal{F}$, define the sample-consistent set
\begin{equation}
\mathcal{H}(D_t;\mathcal{F})
=
\{f\in\mathcal{F}: f(x_i)=y_i \;\forall (x_i,y_i)\in D_t\}.
\end{equation}
In our controlled tasks, $\mathcal{H}(D_t;\mathcal{F})$ remains nontrivial, so point accuracy and ambiguity-awareness need not coincide.

\subsection{Protocol-conditioned surrogate belief}
Fix a query set $\mathbf{x}=(x_1,\ldots,x_m)$ with $m$ queried locations, and a prompt specification $(p,\pi)$, where $p$ is prompt text and $\pi$ is query protocol. The elicited predictive law over the corresponding value vector $\mathbf{y}\in\mathbb{R}^m$ is
\begin{equation}
B_{\theta}^{p,\pi}(D_t;\mathbf{x})
:=
P_{\theta}(\mathbf{y}\mid \mathbf{x},D_t,p,\pi).
\label{eq:belief-object}
\end{equation}

\begin{definition}[Protocol-conditioned surrogate belief]
\label{def:belief}
For fixed $(D_t,p,\pi)$, the family
\[
\{B_{\theta}^{p,\pi}(D_t;\mathbf{x}) : m\ge 1,\ \mathbf{x}\in\mathcal{X}^m\}
\]
is the \emph{protocol-conditioned surrogate belief} exposed by the LLM. Here ``belief'' is behavioral: it is the predictive object seen by a downstream optimizer, not a claim about internal latent state.
\end{definition}

Under \pw{}, queries are answered independently:
\begin{equation}
B_{\theta}^{p,\pw}(D_t;\mathbf{x})
=
\prod_{j=1}^{m} P_{\theta}(y_j\mid x_j,D_t,p,\pw).
\label{eq:pw-factorization}
\end{equation}
Under \joint{}, the full query list is answered in one autoregressive completion:
\begin{equation}
B_{\theta}^{p,\joint}(D_t;\mathbf{x})
=
P_{\theta}(\mathbf{y}\mid \mathbf{x},D_t,p,\joint)
=
\prod_{j=1}^{m} P_{\theta}(y_j\mid y_{<j},\mathbf{x},D_t,p,\joint).
\label{eq:joint-factorization}
\end{equation}

\begin{definition}[Belief shift at fixed evidence]
\label{def:belief-shift}
For two prompt specifications $(p,\pi)$ and $(p',\pi')$, and any divergence $\mathsf{D}$ between distributions on $\mathbb{R}^m$, define
\begin{equation}
\Delta_{\mathsf{D}}(\mathbf{x};D_t,(p,\pi),(p',\pi'))
=
\mathsf{D}\!\left(
B_{\theta}^{p,\pi}(D_t;\mathbf{x}),
B_{\theta}^{p',\pi'}(D_t;\mathbf{x})
\right).
\label{eq:belief-shift}
\end{equation}
A nonzero value means that the elicited surrogate changes although the observations do not.
\end{definition}

\begin{remark}[Effective prior, operationally]
\label{rem:prior}
We use \emph{effective prior} to mean systematic belief shift at fixed $D_t$; this is a behavioral rather than Bayesian claim.
\end{remark}

For $Y=(Y_1,\ldots,Y_m)\sim B_{\theta}^{p,\pi}(D_t;\mathbf{x})$ and any linear statistic $S_w=\sum_{j=1}^m w_jY_j$ with weights $w_j\in\mathbb{R}$,
\begin{equation}
\mathrm{Var}(S_w)
=
\sum_{j=1}^m w_j^2\,\mathrm{Var}(Y_j)
+
2\sum_{j<k} w_jw_k\,\mathrm{Cov}(Y_j,Y_k).
\label{eq:curve-var}
\end{equation}
Thus two protocols can agree on pointwise uncertainty yet differ on curve-level uncertainty through cross-location dependence.

\subsection{Prompt axes and study design}
\label{sec:prompt-axes}
% We vary three prompt dimensions: \textbf{query protocol} (\pw{}, which answers each query independently, or \joint{}, which answers all queried locations in a single autoregressive completion), \textbf{prompt style} (a neutral baseline, an \underdet{} warning that states that the observations do not identify a unique function, or an unrelated warning control), and \textbf{structural information} (\texttt{Unknown}, which gives no type information; \texttt{Tell non-linear}, which only states that the function is nonlinear; \texttt{Tell type}, which specifies the function family; \texttt{Tell type + structure}, which adds a qualitative structural property; \texttt{Tell type + parameters}, which adds instance-level parameter information; or \texttt{Wrong type}, which provides an intentionally incorrect family label). Representative templates are given in Appendix~\ref{app:prompts}. These manipulations are designed to isolate mechanism rather than maximize benchmark performance. 

We vary three prompt dimensions: \textbf{query protocol} (\pw{}, which answers each query independently, or \joint{}, which answers all queried locations in a single autoregressive completion), \textbf{prompt style} (a neutral baseline, an \underdet{} warning that states that the observations do not identify a unique function, or an unrelated warning control), and \textbf{structural information} (ranging from no objective information to increasingly specific function family, structural, and parameter cues, as well as intentionally incorrect information). Representative templates are given in Appendix~\ref{app:prompts}. 

The function families used in our experiments---Gaussian, quadratic, linear, sinusoidal, and logistic---are chosen for interpretability and control, rather than as a uniquely privileged benchmark. Similarly, the model suite is designed to span diverse sizes and families, including Llama, GPT, and Qwen variants, covering open- and closed-weight as well as uni- and multimodal models (see Appendix~\ref{app:models} for the full list). Importantly, our elicited beliefs target intuitive (System 1–like) judgments rather than deliberative (System 2–like) reasoning, without eliciting chain-of-thought.
% The function families used in our experiments—Gaussian, quadratic, linear, sinusoidal, and logistic—are selected for interpretability and control, not as a privileged benchmark. Our model suite likewise spans diverse sizes and families, including Llama, GPT, and Qwen variants, across open- and closed-weight as well as uni- and multimodal models; see Appendix~\ref{app:models}. Finally, we elicit intuitive, System 1–like beliefs rather than deliberative System 2 reasoning, without prompting for chain-of-thought.

\subsection{Model-side uncertainty proxies}
\label{sec:belief_uncertainty}
% \paragraph{Token-level uncertainty.}
% For a realized prediction $\hat y(x)$, we use greedy decoding and compute uncertainty
% \begin{equation}
% \mathrm{NLL}_{\theta}(\hat y(x))
% =
% -\log P_{\theta}(\hat y(x)\mid x,D_t,p,\pi),
% \end{equation}
% with confidence transform
% \begin{equation}
% \label{token_conf}
% c_{\theta}^{\mathrm{tok}}(x)=\exp\!\bigl(-\mathrm{NLL}_{\theta}(\hat y(x))\bigr).
% \end{equation}

\paragraph{Token-level uncertainty.}
For a greedy numeric prediction $\hat y(x)$ with output tokens $s_{1:L}$, we use
\begin{equation}
\mathrm{NLL}_{\theta}^{\mathrm{tok}}(\hat y(x))
=
-\frac{1}{L}\sum_{\ell=1}^{L}
\log P_{\theta}(s_\ell\mid s_{<\ell},x,D_t,p,\pi),
\end{equation}
with confidence proxy
$c_{\theta}^{\mathrm{tok}}(x)=\exp(-\mathrm{NLL}_{\theta}^{\mathrm{tok}}(\hat y(x)))$.

\paragraph{Sampling-based uncertainty.}
For output-level uncertainty, we repeatedly sample under fixed $(x,D_t,p,\pi)$ at temperature $=1$ and define
\begin{equation}
u_{\theta}^{\mathrm{sam}}(x)
=
\operatorname{Var}\!\left(\hat y^{(1)}(x),\ldots,\hat y^{(n)}(x)\right).
\end{equation}

\subsection{Reference ambiguity and uncertainty alignment}
\label{sec:alignment-setup}
Realized error alone is insufficient under underdetermination. We therefore introduce a reference ambiguity profile
\begin{equation}
\mcvar(x;D_t)
=
\operatorname{Var}_{f\sim q_D}[f(x)],
\label{eq:vh_main}
\end{equation}
where $q_D$ is a reference distribution over functions that remain sample-consistent with $D_t$.

For a controlled parametric family $f_\phi$ with parameter $\phi\in\Phi$, we define
\begin{equation}
q_D(\phi)
\propto
p_0(\phi)\,
\mathbf{1}\!\left[
\max_{(x_i,y_i)\in D_t}|f_\phi(x_i)-y_i|
\le
\varepsilon_{\mathrm{cons}}
\right],
\label{eq:qD_main}
\end{equation}
where $p_0$ is a broad proposal distribution and $\varepsilon_{\mathrm{cons}}$ is the consistency tolerance. We approximate $\mcvar$ by Monte Carlo on a fixed query grid (Appendix~\ref{app:mcvar}); it is a reference signal for residual ambiguity, not a claim about true posterior variance.

\begin{definition}[Uncertainty alignment]
\label{def:alignment}
For a query grid $\mathbf{x}=(x_1,\ldots,x_m)$ and a pointwise model-side uncertainty proxy $u(x)$, define
\begin{equation}
A(u;D_t,\mathbf{x})
=
\rho\!\left(
(u(x_j))_{j=1}^{m},
(\mcvar(x_j;D_t))_{j=1}^{m}
\right),
\label{eq:alignment}
\end{equation}
where $\rho$ denotes Spearman rank correlation.
\end{definition}

We report $A(u_{\theta}^{\mathrm{sam}};D_t,\mathbf{x})$ and the analogous token-level version using $\mathrm{NLL}_{\theta}(\hat y(x))$. This asks whether more ambiguous regions are assigned higher uncertainty.

A useful idealized baseline is the width functional
\begin{equation}
W_{\mathcal H}(x;D_t)
=
\sup_{f,g\in\mathcal H(D_t;\mathcal F)} |f(x)-g(x)|.
\end{equation}

\begin{proposition}[Consistency-set contraction]
\label{prop:contraction}
If $D_t\subset D_{t+1}$, then $\mathcal H(D_{t+1};\mathcal F)\subseteq \mathcal H(D_t;\mathcal F)$, and hence
\begin{equation}
W_{\mathcal H}(x;D_{t+1})
\le
W_{\mathcal H}(x;D_t)
\qquad
\forall x\in\mathcal X.
\end{equation}
\end{proposition}

\begin{remark}[Sequential implication]
\label{rem:update}
Exact sample-consistent ambiguity cannot expand as new observations are added. Any temporary rise in model uncertainty or drop in confidence under sequential evidence therefore reflects model-side belief revision, not feasible-set expansion alone.
\end{remark}

\section{Static Belief under Underdetermination} 
\label{sec:static} 

Unlike analytic surrogates, where priors must be built into kernels, architectures, or features, LLM surrogates can receive soft structural information in natural language. We ask whether this verbal structure changes optimizer-facing predictions, rather than merely eliciting textual acknowledgment.

\subsection{Prompt structure changes the implied prior} 
\label{sec:static-structure} 

\begin{figure*}[h] \centering \includegraphics[width=1.0\textwidth]{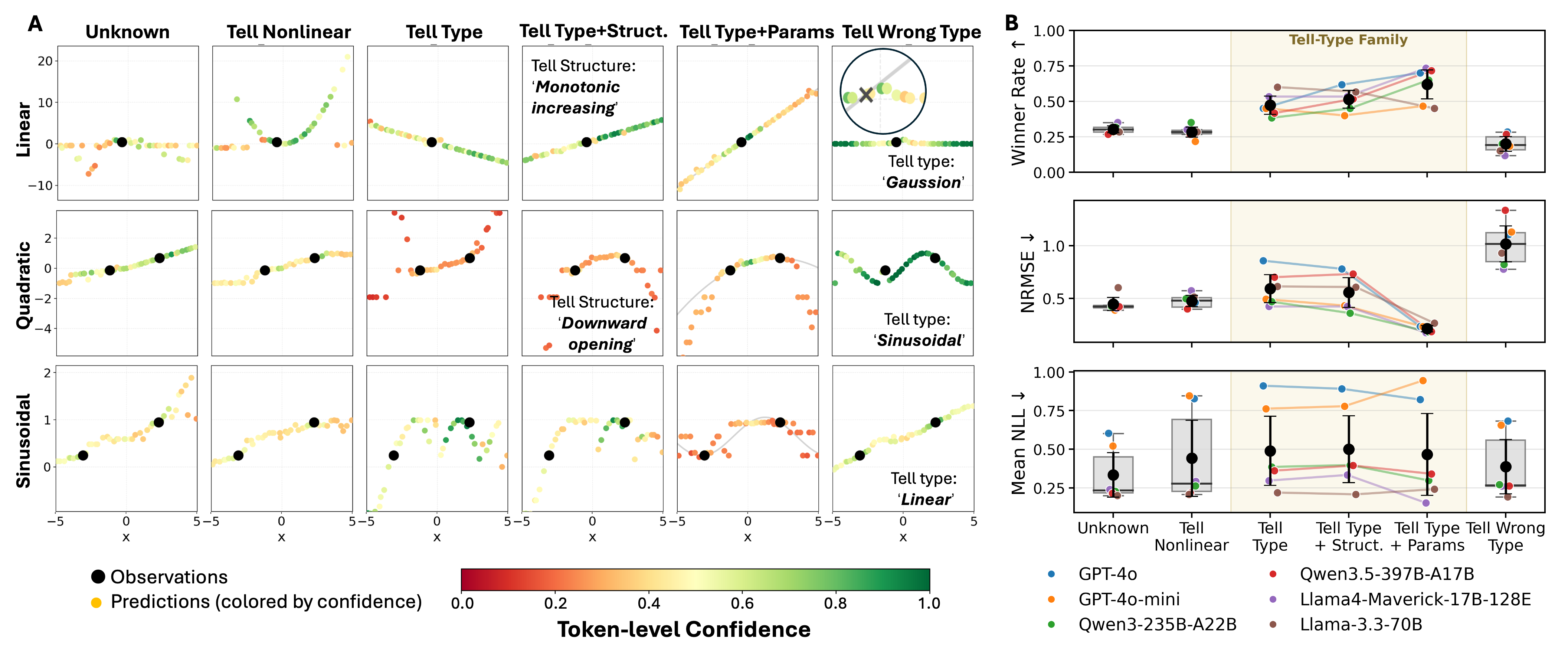} \caption{\textbf{A:} GPT-4o pointwise completions. \textbf{B:} Aggregate prompt effects across models.}
\label{fig:static-structure-grid} \end{figure*}

Holding \(D_t\) fixed and varying only the structural language in \(p\) isolates a prompt-induced belief shift in the sense of Definition~\ref{def:belief-shift}. Under \pw{} querying, Figure~\ref{fig:static-structure-grid}A shows that, with sparse observations, correct structural descriptions pull predictions toward the intended family, whereas wrong-type descriptions push them away. Prompt language therefore acts as an effective prior over completions in the sense of Remark~\ref{rem:prior}. The figure also reveals that independently elicited predictions surprisingly remain coherent across query locations. We further discuss this tendency in Section~\ref{sec:protocol}.

% To summarize across models and 4 function families, we use two metrics on the elicited curve: \emph{winner rate}, based on whether the predicted curve is assigned to the correct function family under AICc-based family selection (Appendix~\ref{app:shape-id}), and NRMSE, which measures pointwise fit quality (Appendix~\ref{app:nrmse}). Figure~\ref{fig:static-structure-grid}B shows the pattern. In the top panel, the \texttt{Tell-Type Family} conditions improve family recovery relative to \texttt{Unknown}, whereas \texttt{Tell Wrong type} degrades it. Within this family, winner rate rises as more correct structural information is added. In the middle panel, the same progression improves NRMSE, indicating closer pointwise fit. The bottom panel shows shifts in mean NLL, so these prompt changes affect uncertainty as well as accuracy. \texttt{Tell Nonlinear} provides only a weak reminder and little fitting-relevant information, as reflected in its small effects on winner rate and NRMSE, yet it still increases uncertainty in some cases. Changes in uncertainty are also model-dependent: larger models such as GPT-4o and Qwen more sharpen as structure accumulates, whereas some smaller models become more diffuse under more complex formula descriptions.

We summarize the elicited curves across models and 4 function families, using two metrics: \emph{winner rate}, the fraction assigned to the correct function family by AICc-based family selection (Appendix~\ref{app:shape-id}), and NRMSE, which measures pointwise fit (Appendix~\ref{app:nrmse}). Figure~\ref{fig:static-structure-grid}B shows that \texttt{Tell-Type Family} prompts improve family recovery and NRMSE relative to \texttt{Unknown}, whereas \texttt{Tell Wrong type} degrades both. Adding more correct structural information further improves recovery within the same family. Prompt changes also affect uncertainty, as reflected by shifts in mean NLL. \texttt{Tell Nonlinear} provides only a weak cue and has little effect on fit, but can still change uncertainty. Changes in uncertainty are also model-dependent: larger models such as GPT-4o and Qwen tend to become more confident as more structural information is provided, whereas some lower-capacity models such as GPT-4o-mini and Llama-3.3-70B become more diffuse under more complex formula descriptions.

Additionally, even without observations, LLM beliefs can show structured, model-dependent trends such as linearity or monotonicity (Appendix~\ref{app:belief_patterns}). These trends can persist far outside the main range \( [-5,5] \), surprisingly even for inputs as large as \(10^1\)--\(10^4\) (Appendix~\ref{app:large-x-extrapolation}). However, LLMs are not always able to verbalize these numerical biases (Appendix~\ref{app:verbalization}). Predicting a single point can often be solved through direct completion, whereas describing the global shape requires integrating a higher-level abstraction. This highlights a key gap: verbal reasoning is not equivalent to the internal computation underlying numerical prediction.

Together, these results support two conclusions: (1) the elicited LLM surrogate is shaped by both sparse numerical observations and prompt language; (2) structural prompts act as effective priors, with correct cues improving fit and wrong-type cues misleading the surrogate.

\subsection{\pw{} and \joint{} are not interchangeable elicitation protocols}
\label{sec:protocol}

\begin{figure*}[h]
    \centering
    \includegraphics[width=0.9\linewidth]{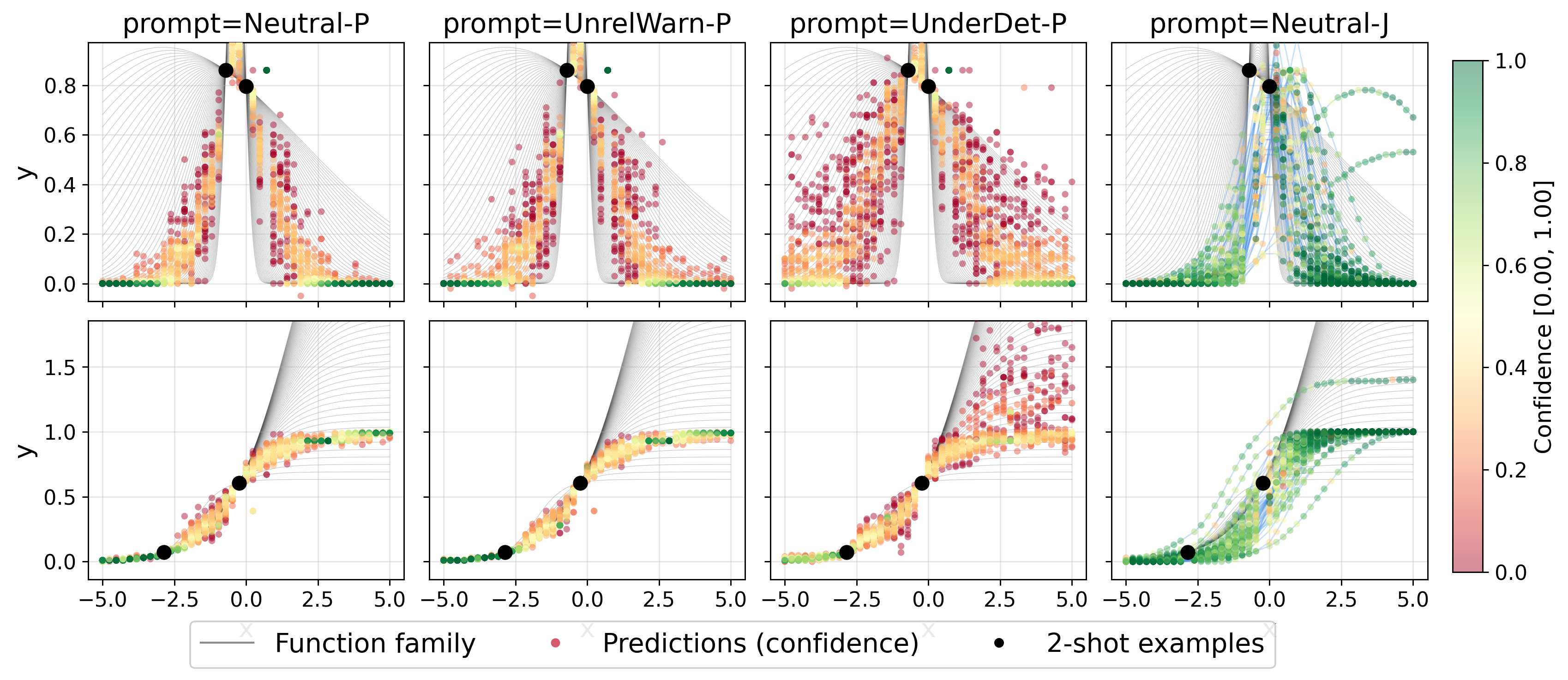}
    \caption{50 repeated completions of the same sparse-observation task in GPT-4o. Under \pw{}, \underdet{} broadens plausible completions while preserving coarse structure. }
    \label{fig:protocol-examples}
\end{figure*}

We next hold $D_t$ fixed and change only the elicitation protocol. The task uses two observations and a partially specified hypothesis class in which the underlying family is Gaussian or logistic. We sample 50 untruncated temperature-1 completions per condition to compare \pw{} and \joint{} with minimal decoding confounds.

In Figure~\ref{fig:protocol-examples}, point color denotes token-level uncertainty and the background shows the sample-consistent set $\mathcal{H}(D_t;\mathcal F)$. Under \pw{}, independently answered locations still organize into coherent curves, revealing a strong default bias (which would otherwise be scattered without such bias). Adding an \underdet{} warning broadens the prediction distribution in more ambiguous regions. Under \joint{}, later values condition on earlier generated values, so completions become more path-dependent and internally coherent, but less faithful to the original observations.
% \footnote{
% LLM believe its bullsh*t. 
% The joint distribution is more self-consistent than the pointwise, but also generally less faithful to the observations (we speculate that this is because the model is attending to its own predictions so far (rather than the original problem).
% and is therefore, in a sense, "beleiving its own BS". 
% }
% \footnote{LLMs may over-rely on their own predictions. The joint distribution is more self-consistent than the pointwise, but also generally less faithful to the observations (we speculate that this is because the model is attending to its own predictions so far rather than the original problem).}
Interestingly, beyond curve-level consistency, we observe additional biases in the Gaussian setting: predictions rarely exceed an apparent upper bound near $1$, and completions often treat the two observed points as lying on opposite sides of a symmetry axis rather than on the same tail.

\begin{figure*}[h]
    \centering
    \includegraphics[width=0.8\textwidth]{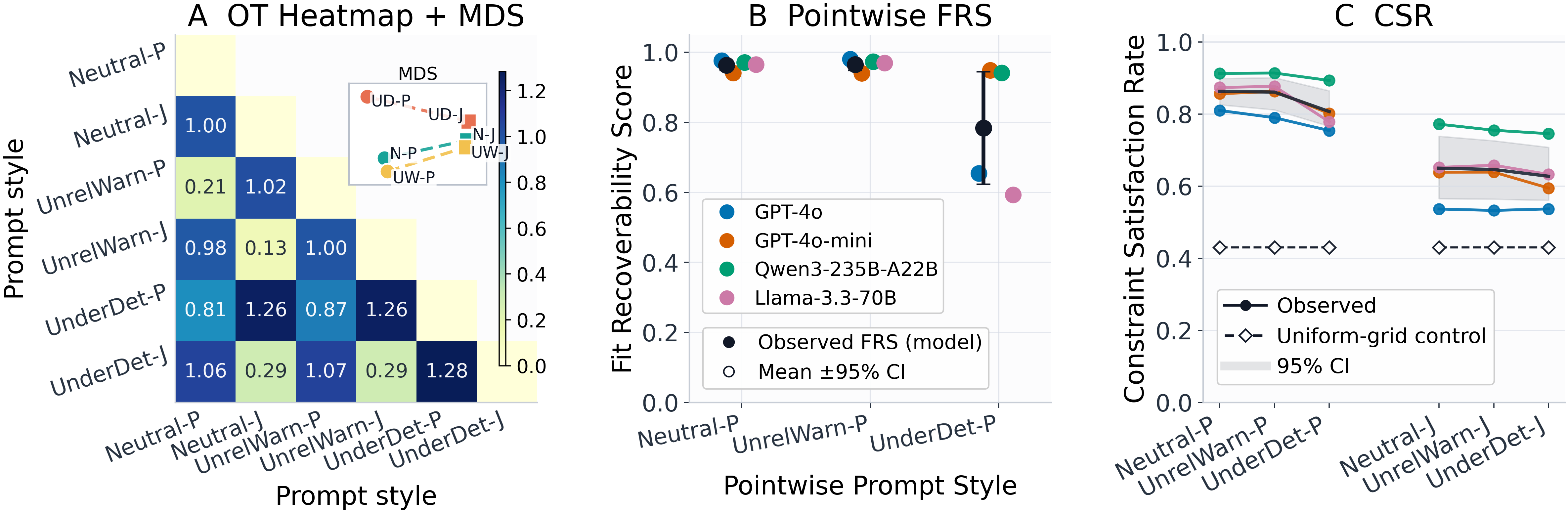}
    \caption{Aggregate effects of prompt style and protocol. 
\textbf{A:} OT distance (with Multidimensional Scaling inset). 
\textbf{B:} fit recoverability score. 
\textbf{C:} constraint satisfaction rate.}
    \label{fig:protocol-summary}
\end{figure*}

Figure~\ref{fig:protocol-summary} quantifies these differences across six models. Panel~A estimates belief shift at fixed evidence via Optimal Transport (OT) distance between sampled completion distributions. Under \pw{}, \underdet{} moves substantially away from neutral while the unrelated-warning control remains close; under \joint{}, these distances shrink, suggesting that it compresses prompt effects. Panel~B reports the pointwise Fit Recoverability Score (FRS; Appendix~\ref{app:frs}), which measures whether repeated predictions can be explained by one consistent fit from the stated family. Panel~C reports the Constraint Satisfaction Rate (CSR; Appendix~\ref{app:csr}), which measures whether predictions are consistent with the stated function family. Under \pw{} \underdet{}, FRS is generally lower, consistent with reduced inductive bias. Under \joint{}, CSR is lower, consistent with autoregressive error propagation. This also explains why token-level and sampling-based uncertainty can diverge: \joint{} outputs may be locally confident yet vary across runs.

% These results support three conclusions: (1) query protocol is part of the surrogate specification, not a formatting detail; \joint{} suppresses prompt effects and weakens constraint fidelity, whereas \pw{} more faithfully exposes prompt-induced variation; (2) under \pw{}, neutral prompting already yields globally coherent predictions, revealing a strong default bias toward coherent functional completion; and (3) within \pw{}, task-relevant underdetermination warnings broaden the completion distribution while largely preserving compatibility with the stated family, though the effect varies across models.

These results support three conclusions: (1) query protocol is part of the LLM surrogate specification, not a formatting detail: \joint{} suppresses prompt effects and weakens constraint fidelity, whereas \pw{} more faithfully exposes prompt-induced variation; (2) neutral \pw{} prompting reveals a bias toward globally coherent completion; (3) task-relevant underdetermination warnings broaden \pw{} completion distribution while largely preserving compatibility with the stated family, although effect sizes vary by model.

\begin{wrapfigure}{r}{0.52\textwidth} 
    \centering
    \vspace{-10pt}
    \includegraphics[width=\linewidth]{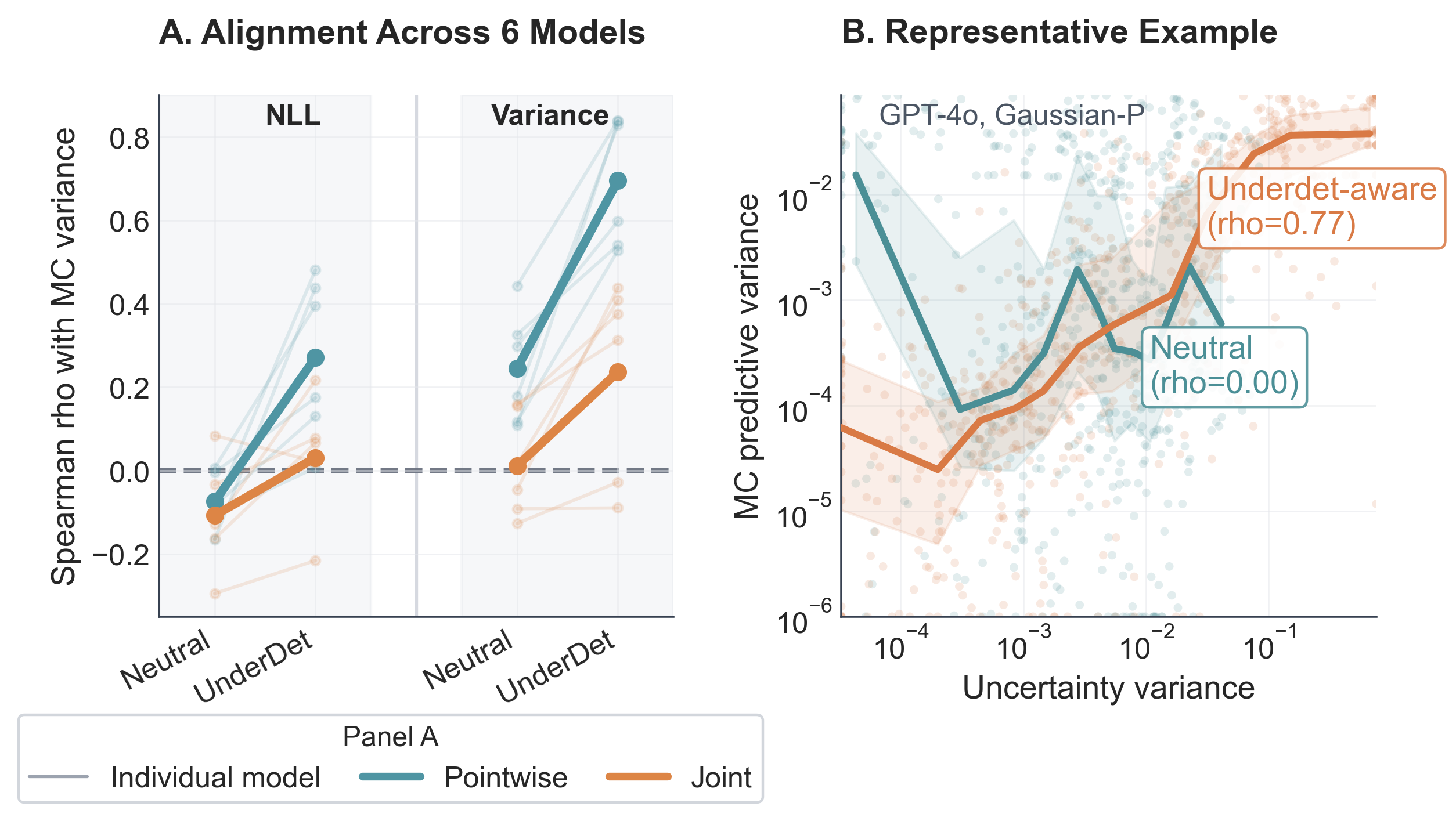}
    \vspace{-3pt}
    \caption{Uncertainty alignment. 
\textbf{A:} Spearman correlation. 
\textbf{B:} Gaussian example (GPT-4o).}
    \label{fig:uncertainty-alignment}

\end{wrapfigure}

\section{Uncertainty Alignment}
\label{sec:alignment}

Standard calibration asks whether confidence tracks realized error. Under underdetermination, this is incomplete: a prediction may be accurate yet overconfident because many alternative functions remain consistent with the observations. Definition~\ref{def:alignment} instead asks whether the model’s uncertainty correctly reflects which locations are inherently more ambiguous. We compute the reference ambiguity profile in Eq.~\eqref{eq:vh_main} using the Monte Carlo sample-consistent construction in Section~\ref{sec:alignment-setup} and Appendix~\ref{app:mcvar}, and compare it with LLM-side uncertainty via Spearman correlation. We use both token level (NLL) and sampling variance as LLM uncertainty proxies. Because the diagnostic is rank-based, these proxies need not share a scale; the question is simply whether more ambiguous regions receive higher uncertainty. 

Figure~\ref{fig:uncertainty-alignment} shows the main pattern. Neutral prompts often give weak or negative alignment. In \pw{} mode, \underdet{}-aware prompting improves both variance- and NLL-based alignment: uncertainty increases in ambiguous regions and decreases in less ambiguous ones. The same intervention is much weaker in \joint{} mode. In both query protocols, the LLM understands the underdetermination warning at a semantic level, but this does not mean it can incorporate it into an optimizer-facing surrogate belief in both cases.

\section{Belief Updating under Sequential Evidence}
\label{sec:update}

Previous sections study LLM surrogate belief under a fixed observation set. Since optimization is sequential, the surrogate should also update as evidence accumulates. We therefore examine how the induced surrogate evolves as evidence is sequentially accumulated under different ordering conditions.
\begin{wrapfigure}{r}{0.6\textwidth} 
    \centering
    \includegraphics[width=\linewidth]{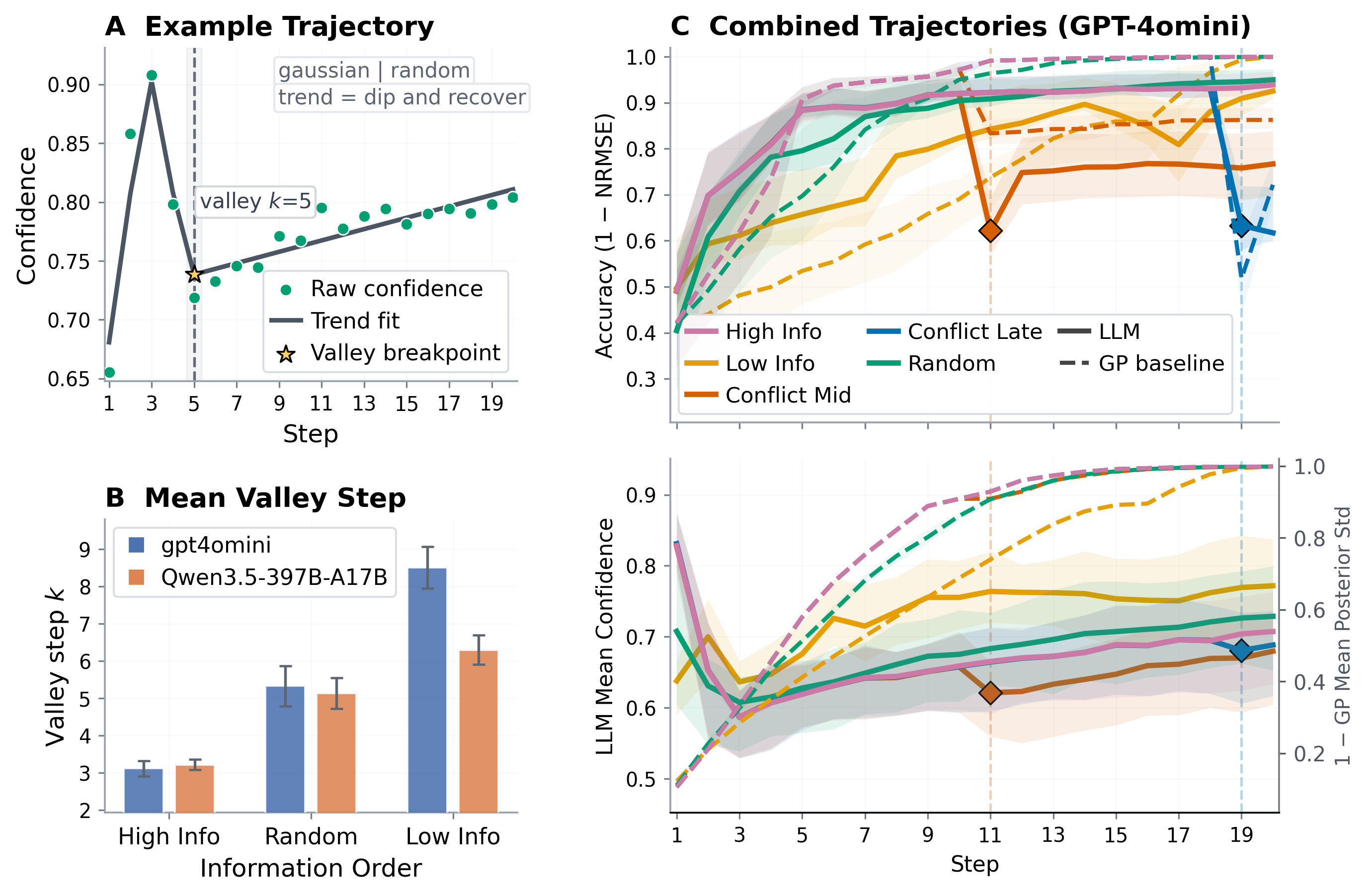}
    \vspace{-3pt}
    \caption{Sequential belief updating. A: Example run. B: Valley step. C: Accuracy/confidence trajectories.}
    \label{fig:sequential-updating}
    \vspace{-3pt}
\end{wrapfigure}
For each ordering $o$, observations form nested prefixes 
$D_1^{(o)}\subset\cdots\subset D_T^{(o)}$. Non-conflict orders reveal the same candidate pool by decreasing information 
(High Info), increasing information (Low Info), or random permutation, where information is quantified by GP-estimated posterior-variance reduction over the query grid. These conditions therefore isolate pure order effects under identical evidence. The conflict orders insert one contradictory point at a midway or late prefix step in the same high-information support sequence, chosen at a GP-confident location and shifted away from the GP prediction (Appendix~\ref{app:sequential-details}). After each prefix, we record accuracy \(1-\mathrm{NRMSE}_t\) and mean token confidence \(\bar c_t\). We summarize the lowest-confidence point by the \emph{confidence valley} step \(t_{\mathrm{valley}}=\operatorname*{arg\,min}_{t}\bar c_t\) (Appendix~\ref{app:metrics}). For comparison with a conventional surrogate, we replay the same prefixes with a fixed-kernel GP surrogate and map uncertainty to confidence by \(1-\sigma_t^{\mathrm{GP}}(x)\).

Figure~\ref{fig:sequential-updating}A shows a dip-and-recovery confidence trajectory rather than monotonic contraction. As an informal analogy, this resembles the Dunning--Kruger effect in psychology \citep{kruger1999unskilled}: individuals may initially overestimate their ability, but confidence drops as expertise increases, producing a \emph{valley of despair}\footnote{The analogy is descriptive only. We use it to name the shape of the confidence trajectory, not to claim that the LLM has human-like metacognition or a psychological self-assessment process.}. As expertise continues to accumulate, confidence then recovers, yielding the observed dip-and-recovery pattern. Figure~\ref{fig:sequential-updating}B shows that the order of evidence shapes when the confidence valley occurs. High Info, where GP-ranked high-information observations are revealed first, moves the valley earlier, consistent with faster revision of the initial hypothesis. Low Info delays the valley, leaving the model confident but weakly updated for longer. Conflict points cause sharper disruption: accuracy drops abruptly and confidence destabilizes, with Conflict Late recovering least because fewer supportive observations remain.

% The non-conflict conditions differ only in reveal order: \texttt{high\_info\_first} presents the most informative remaining points early, \texttt{low\_info\_first} presents the least informative points early, and \texttt{random} uses a random permutation. These conditions therefore isolate pure order effects under identical evidence. 
% The non-conflict orders reveal the same observation pool by decreasing information, increasing information, or random permutation, isolating order effects under fixed evidence.
% The conflict conditions probe a different regime. \texttt{conflict\_middle} and \texttt{conflict\_late} follow the same high-information support sequence but insert a single contradictory observation midway or near the end. The conflict point is chosen where the current GP reference is most confident, and its \(y\)-value is shifted away from the current prediction to make the new evidence maximally surprising relative to the prefix. Full details are given in Appendix~\ref{app:sequential-details

% After each update, we re-query the model on a fixed evaluation grid and track accuracy and mean confidence. Accuracy at step \(t\) is defined as \(1-\mathrm{NRMSE}_t\), and mean confidence is the average of token-level confidence \(c_{\theta}(x;D_t^{(o)},p,\pi)\) over \(x \in \mathcal{X}_{\mathrm{qry}}\), with \(\pi=\pw{}\). 
% To separate model-side revision from observation-sequence geometry, we replay the same prefixes with a fixed-kernel GP control and compute the same summaries, converting GP uncertainty to confidence as \(1-\sigma_t^{\mathrm{GP}}(x)\).

The GP control tracks evidence geometry more directly: uncertainty changes monotonically, contracting earlier under \texttt{high\_info\_first}, more gradually under \texttt{low\_info\_first}, and being locally perturbed by the inserted conflict point. The stronger non-monotonicity in the LLM therefore reflects model-side belief revision rather than evidence geometry alone. Unlike the GP, which reaches the same final state for a fixed observation set, the LLM remains order-sensitive and can produce different outcomes under different evidence orders.

Overall, the LLM surrogate updates as more evidence is provided, but not solely as a function of the final observation set. Since exact consistency would only reduce residual ambiguity as observations accumulate, the non-monotonic confidence curves in Figure~\ref{fig:sequential-updating} indicate internal belief revision rather than simple feasible-set narrowing.

\section{Do Elicited Belief Differences Affect BO Decisions?}
\label{sec:bo}

Previous sections reveal differences in elicited surrogate beliefs, but not whether they affect optimization behavior. We therefore use the LLM predictions as BO surrogates and measure their effect. These experiments are not general-purpose optimizer benchmarks; instead, they test whether belief differences persist under BO decisions. Acquisition scores are computed from numeric LLM predictions and token-level uncertainty: UCB-style ranking score for Branin and an EI-style ranking score for HPO and battery design. The uncertainty is used as a heuristic exploration signal, not as a calibrated objective-space predictive standard deviation. The three studies are designed to cover increasingly realistic settings: a controlled benchmark with deliberate prior mismatch, a practical HPO task with limited domain structure, and a battery-design task where the prompt can provide useful scientific context. All BO studies use $10$ seeds per condition, and all LLM calls use direct numeric prediction without chain-of-thought. The LLMs are GPT-4o for Branin, Qwen3.5-397B-A17B for HPO, and GPT-5.4 for battery design. These studies are not intended as cross-model comparisons. For each task, we use a representative LLM and compare prompt/protocol variants within the same task. Additional experimental details and prompts are provided in Appendix~\ref{app:details}.

\textbf{Controlled prior-mismatch benchmark.}
We first run a small BO loop on canonical Branin and on a modified Branin variant. In both cases, the objective is described only as an unknown function. The modified variant preserves the Branin-like shape but moves the optima away from the canonical locations; the position of the initial observations is also chosen to have values similar to canonical Branin. This creates a setting in which a familiar Branin prior can be useful on the canonical task but misleading when the optimum is displaced.

\begin{figure*}[h]
    \centering
    \includegraphics[width=0.85\textwidth]{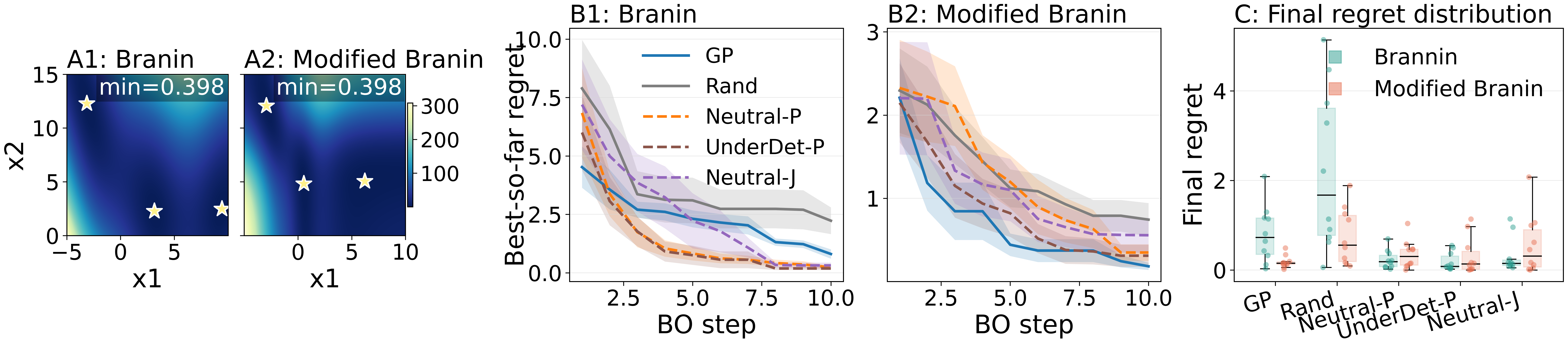}
\caption{Branin BO study. A: Objectives. B: Mean best-so-far regret. C: Final regret.}
    \label{fig:branin-bo}
\end{figure*}

\begin{wrapfigure}[11]{r}{0.23\textwidth}
    \vspace{-10pt}
    \centering
    \includegraphics[width=\linewidth]{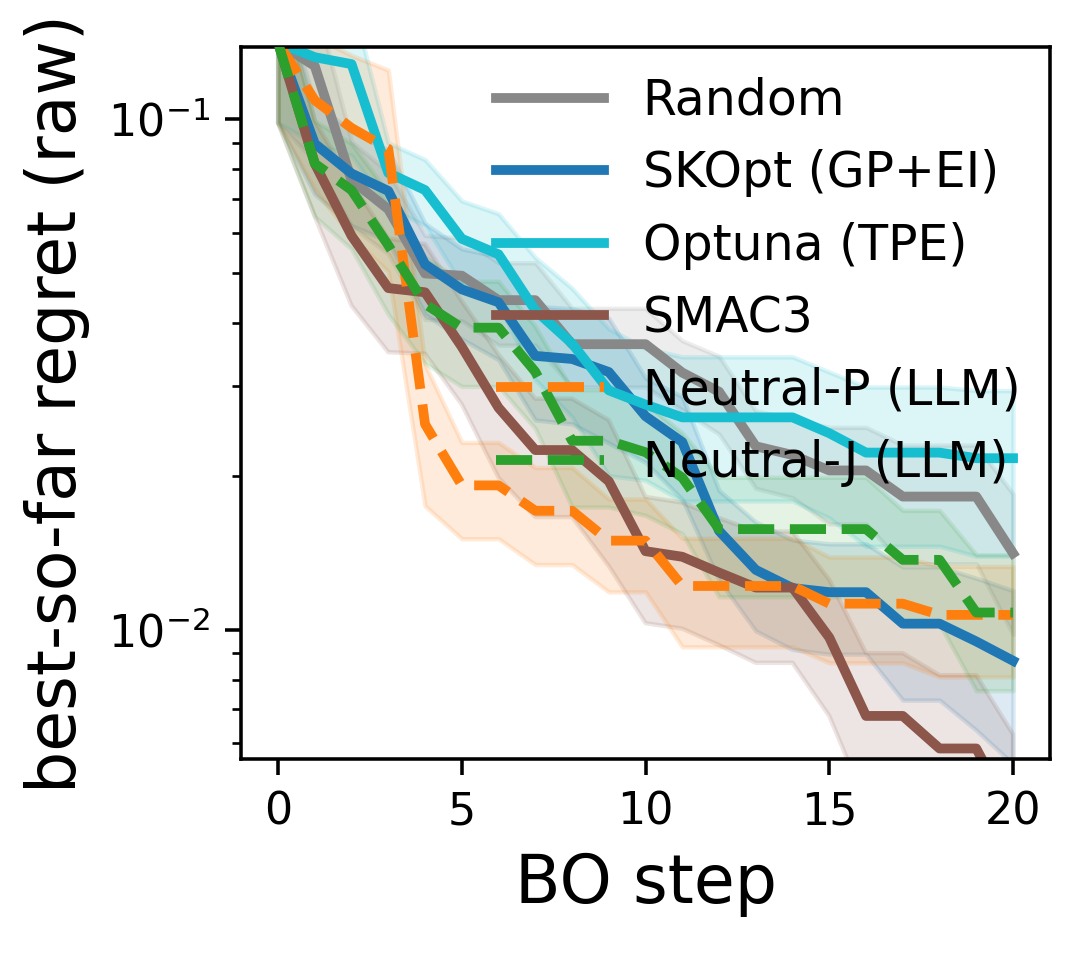}
    \vspace{-10pt}
    \caption{Mean best-so-far regret (HPO).}
    \label{fig:hpo}
\end{wrapfigure}

Figure~\ref{fig:branin-bo} shows the effect of prior mismatch. On canonical Branin, LLM \pw{} achieve lower regret than the GP baseline, but this advantage disappears once the optima are modified, consistent with the fact that the LLM prior influences BO decisions. POINTWISE outperforms JOINT, and the underdetermination-aware prompt further helps on the modified task, matching the static diagnostics: POINTWISE better fits the observations, while the underdetermination prompt improves alignment. Thus, the prompt and protocol effects observed statically also lead to different BO decisions.

\textbf{HPO benchmark.}
We next test whether these protocol effects persist in a more practical setting: low-budget XGBoost hyperparameter optimization on California Housing. The LLM receives model family and task type, but not the dataset name or any dataset-specific semantic description. Figure~\ref{fig:hpo} shows that POINTWISE reduces regret fastest in early stages, while JOINT is weaker for much of the trajectory. SMAC3 achieves the best final regret. These results suggest that the POINTWISE--JOINT difference is not limited to toy functions, and that LLM priors can drive rapid early-stage progress even when only weak task structure is provided.

\begin{wrapfigure}[11]{r}{0.23\textwidth} 
    \vspace{-12pt}
    \centering
    \includegraphics[width=\linewidth]{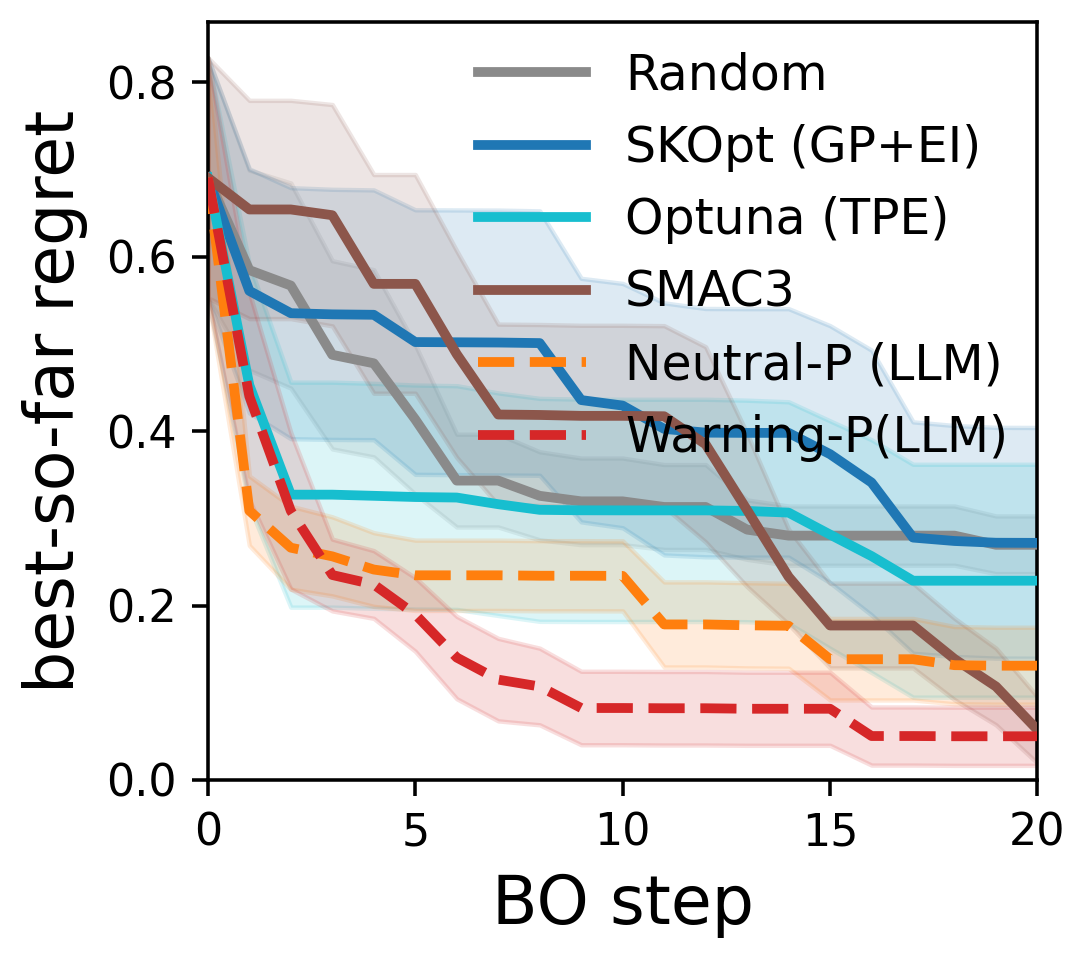}

    \caption{Mean best-so-far regret (battery).}
    \label{fig:battery}

\end{wrapfigure}

\textbf{Domain-structured battery design.}
Finally, we evaluate a domain-structured cathode-design task: optimizing positive-electrode thickness and porosity using DFN-simulated energy densities at 0.5C and 3C. The objective rewards high energy density at both rates while penalizing imbalance. Unlike black-box baselines, the LLM receives indirect domain information through variable names, battery context, and the objective definition; encoding this equivalently in classical optimizers would require task-specific surrogate design and is not always well-defined. The warning prompt further emphasizes broad domain guidance (the LLM may already possess): low-rate energy density improves with lower inactive content, while high-rate performance depends more strongly on transport pathways and porosity. But it does not reveal the simulator response surface or good designs. Figure~\ref{fig:battery} shows the largest benefit for LLM surrogates. Both POINTWISE prompts achieve the best optimization performance, rapidly identifying high-scoring regions. The warning prompt further improves performance, suggesting that informative domain guidance affects downstream optimization decisions. 

Across the BO studies, we use optimization as a test of elicited beliefs rather than as a broad optimizer comparison. The Branin study shows that LLM priors can help under alignment but hurt under mismatch. The HPO study shows that the POINTWISE--JOINT gap persists beyond synthetic objectives. The battery task shows that LLM surrogates are most useful when the prompt exposes domain structure that generic black-box baselines cannot access. Together, these results show that prompt and protocol affect BO decisions, and should be treated as part of the LLM surrogate specification. LLMs also make BO more accessible through language-specified priors.

\section{Related Work}
\label{sec:related}

\textbf{In-context learning and surrogate regression.}
A growing literature views in-context learning as inference-time adaptation
\citep{brown2020language,garg2022can}. Closely related, \citet{requeima2024llmprocesses}
introduce LLM Processes, which elicit numerical predictive distributions from
LLMs conditioned on natural language. Our work is complementary: we study the
optimizer-facing surrogate belief elicited from an LLM, focusing on how this
belief depends on observations, prompt text, and query protocol. Beyond predictions,
we analyze uncertainty under static evidence and sequential evidence
updates.

\textbf{LLMs for Bayesian optimization.}
LLMs have also been used to improve Bayesian optimization \citep{liu2024large,snoek2012practical,shahriari2016taking}. These works primarily ask whether LLMs can improve optimization performance. By contrast, we hold the broad optimization setup fixed and ask how prompt text and query protocol affect the surrogate belief used by the acquisition function.

\textbf{Mechanistic views of in-context learning.}
A parallel literature studies the mechanisms underlying ICL. Some interpret it as implicit Bayesian inference over latent task structure \citep{xie2022explanation}, while others show that transformers can implement simple learning algorithms, such as linear regression or gradient-based updates, in the forward pass \citep{garg2022can,akyurek2022learning,vonoswald2023transformers}. Mechanistic interpretability work further links ICL to circuit-level components such as induction heads \citep{olsson2022context}. We treat LLMs as distinct from these mechanisms, even if they may exhibit similar behaviors. We do not adjudicate among these mechanisms. Our use of ``belief'' is behavioral: it refers to the finite-dimensional predictive law elicited from a fixed LLM. 

\textbf{Uncertainty and calibration.}
Calibration \citep{guo2017calibration,kuleshov2018accurate,ovadia2019can} has recently
been studied for LLM uncertainty
\citep{jiang2021know,kuhn2023semantic}. These works typically evaluate
uncertainty against realized error. In sparse-observation, however,
accurate predictions can still be overconfident when many functions remain
consistent with the data. We therefore study \emph{uncertainty alignment}: whether
LLM uncertainty tracks residual function-space ambiguity. Related work finds that multi-turn LLM confidence need not improve monotonically as evidence accumulates \citep{li2024can}. We further show that the resulting confidence valley is order-sensitive, with its location depending on the sequence in which evidence is revealed.

\section{Discussion and Conclusions}
\label{sec:discussion}

% This paper studies how an LLM-based surrogate behaves under sparse observations when the prompt text and query protocol are varied. The main contribution is a formal measurement lens for this setting, not a claim that LLMs implement principled Bayesian inference. The core object is the \emph{protocol-conditioned surrogate belief}: the optimizer-facing predictive law induced by a given combination of observations, prompt text, and query protocol. Prompt-induced prior shift, structural recovery, constraint fidelity, uncertainty alignment, and sequential revision are all different projections of this same elicited object.

% From this perspective, uncertainty alignment is best understood as a diagnostic for controlled settings: it separates residual ambiguity in the sample-consistent hypothesis space from realized prediction error, without claiming to recover the model's true posterior variance or to provide a general deployment-time uncertainty measure. The same lens clarifies why prompt language matters beyond verbal fluency, since a model may reproduce task-relevant terms without using them to revise the surrogate that an optimizer actually interacts with. Our baseline set is therefore intentionally small: the goal is not broad optimizer comparison, but to isolate how prompting and query protocol change the behavior of the same LLM-based surrogate under otherwise fixed observations.

This paper studies how prompt text and query protocol affect the behavior of an LLM-based surrogate under sparse observations. The central object is the \emph{protocol-conditioned surrogate belief}: the optimizer-facing predictive law induced by observations, prompt text, and query protocol. It treats uncertainty alignment, structural recovery, constraint fidelity, sequential revision, and BO regret as different projections of the same elicited object and asks whether prompt language actually changes the surrogate seen by the optimizer rather than merely being verbally reproduced. 
Our controlled experiments show that LLM surrogates are shaped not only by data, but also by how they are elicited. Correct structural language acts as an effective prior, while wrong-type language can systematically mislead the surrogate. \pw{} and \joint{} are not interchangeable interfaces: \pw{} better preserves ambiguity-sensitive uncertainty and observation faithfulness, whereas \joint{} yields coherent but less constraint-faithful completions. Under sequential evidence, the surrogate revises in non-monotonic and order-sensitive ways. In BO, these differences affect acquisition behavior and regret. In LLM-based BO, elicitation protocol is therefore not an implementation detail; it changes the finite-dimensional predictive law used by the acquisition function.

% Overall, our controlled experiments show that sparse-observation LLM surrogates are shaped not only by the data, but also by how they are elicited. Prompt acts as an effective prior in the operational sense of Remark~\ref{rem:prior}: correct structural cues steer completions toward the appropriate function family, while wrong-type cues can systematically mislead the surrogate. \pw{} and \joint{} are not interchangeable querying interfaces: \pw{} better preserves ambiguity-sensitive behavior and uncertainty alignment, whereas \joint{} produces coherent but less observation-faithful completions. Under sequential evidence, the induced surrogate revises in non-monotonic and order-sensitive ways, and contradictory observations can sharply disrupt confidence and accuracy. These elicited differences also matter downstream: in our BO study, pointwise querying is generally more reliable than joint querying, and underdetermination-aware prompting is especially helpful when
% prior mismatch. The results also suggest
% that LLM surrogates can meaningfully alter BO behavior when their priors are informative or when prompts elicit useful information.

\textbf{Broader Impact.} In real low-data optimization settings such as robotic-lab control, decisions must integrate sparse observations with safety limits, resource constraints, machine precision, and human expectations about plausible values. The practical question is how much the elicited surrogate should rely on LLM prior knowledge and observed evidence. Prompt design and query protocol are therefore part of the LLM surrogate specification because they shape the balance between prior-driven completion, observation faithfulness, and global coherence in downstream decisions. 
\textbf{Limitations.}
The study is intentionally controlled: the tasks are mostly noise-free, and the ambiguity reference is defined only for families where a sample-consistent profile can be approximated. This improves interpretability, but leaves open how the same effects scale to noisier settings.

% \textbf{Limitations.}
% The study is intentionally controlled: the tasks are low-dimensional and mostly noise-free, and the ambiguity reference is defined only where a sample-consistent profile can be approximated. This improves interpretability, but leaves open how the same effects extend to noisier and higher-dimensional settings. \textbf{Broader Impact.}

{\small
\bibliographystyle{plainnat}
\bibliography{references}
}

%%%%%%%%%%%%%%%%%%%%%%%%%%%%%%%%%%%%%%%%%%%%%%%%%%%%%%%%%%%%
\newpage
\appendix
\renewcommand{\thefigure}{\Alph{section}.\arabic{figure}}
\setcounter{figure}{0}

\renewcommand{\thetable}{\Alph{section}.\arabic{table}}
\setcounter{table}{0}

\section{Prompt Templates and Experimental Inputs}
\label{app:prompts}

This appendix documents the experimental inputs used throughout the paper. The purpose of these prompt manipulations is diagnostic: they are designed to isolate which aspects of prompt text and query protocol change the optimizer-facing surrogate belief, i.e., the finite-dimensional predictive law defined in the main text.

\subsection{Base task framing}
We use two base prompt formats corresponding to the two query protocols studied in the paper.

\paragraph{Base prompt: pointwise (\pw).}
In the pointwise setting, each query location is asked in a separate completion. A representative template is:
\begin{quote}
\small
You are a function approximator. Given examples of $(x, y)$ pairs from a function, predict the $y$ value for a new $x$.

Rules:
\begin{enumerate}[leftmargin=*]
    \item Output only the numeric $y$ value.
    \item Use the same precision as the examples.
    \item Do not include explanation or extra text.
\end{enumerate}

Here are example points from an unknown function:

\texttt{[1]} $x=x_1$ $y=y_1$ \\
\texttt{[2]} $x=x_2$ $y=y_2$ \\
\ldots

Predict the $y$ value for:

$x=x^{\mathrm{query}}$ \\
$y=$
\end{quote}

\paragraph{Base prompt: joint (\joint).}
In the joint setting, the full query list is answered in a single completion. A representative template is:
\begin{quote}
\small
You are a function approximator. Return strict JSON only. Output must contain exactly one key ``y'' with a list of numeric values. The list order must match the query list order.

Here are example points from an unknown function:

\texttt{[1]} $x=-1.00$ $y=0.20$ \\
\texttt{[2]} $x=0.00$ $y=1.00$ \\
\texttt{[3]} $x=1.00$ $y=0.25$

Predict $y$ for each query in order.

QUERY LIST: \\
\texttt{[1]} $x=-0.50$ \\
\texttt{[2]} $x=0.50$ \\
\texttt{[3]} $x=1.50$ \\
\texttt{[4]} $x=-1.50$

Return: \texttt{\{"y":[y1,...,y4]\}}
\end{quote}

The pointwise format is repeated independently for each query point on the evaluation grid. The joint format produces one vector-valued completion per repeated sample.

\subsection{Prompt style conditions}
We study three prompt styles.

\paragraph{Neutral baseline.}
The neutral condition presents the task directly, with no additional statement about ambiguity.

\paragraph{Underdetermination-aware warning.}
This condition makes the ambiguity of the sparse observation set explicit. A representative sentence is:
\begin{quote}
\small
Warning: this description is underdetermined. With only a few points, many functions of the stated family can still fit the observations.
\end{quote}
Its purpose is not to instruct the model toward any specific function, but to test whether making residual ambiguity explicit changes the predictive distribution or uncertainty.

\paragraph{Unrelated warning control.}
This control preserves the presence of an extra warning-like sentence while removing its task relevance. A representative sentence is:
\begin{quote}
\small
Warning: descriptive details such as color, texture, and smell can matter when discussing food.
\end{quote}
This distinguishes task-relevant ambiguity effects from generic instruction-loading effects.

\subsection{Structural-information conditions}
\label{app:prompt-structure}
We vary the amount and correctness of verbal structural information supplied in the prompt, while keeping the observed input-output examples fixed. The prompt styles are:

\paragraph{Unknown.}
The model is explicitly told that the function type is unknown, and must infer the pattern only from the examples:
\begin{quote}
\small
IMPORTANT: The underlying function type is unknown.
\end{quote}

\paragraph{Tell non-linear.}
The model receives only a coarse shape-level cue, without the function family or parameters:
\begin{quote}
\small
IMPORTANT: The underlying function is nonlinear, not a straight line.
\end{quote}

\paragraph{Tell type.}
The model is told the true function family and its generic formula family, but not the instance-specific parameters:
\begin{quote}
\small
IMPORTANT: The underlying function is \{type\_desc\}.\\
Formula family: \{formula\}.
\end{quote}

\paragraph{Tell type + structure.}
The model is told the true function family, its generic formula family, and one qualitative structural hint derived from the sampled instance:
\begin{quote}
\small
IMPORTANT: The underlying function is \{type\_desc\}.\\
Formula family: \{formula\}.\\
\{structural\_hint\}
\end{quote}

\paragraph{Tell type + parameters.}
The model is told the true function family, the generic formula family, and the numerical parameters of the sampled instance:
\begin{quote}
\small
IMPORTANT: Here are example points from a \{type\_desc\} function.\\
The exact formula is \{formula\}. Parameters: \{params\_str\}.
\end{quote}

\paragraph{Wrong type.}
The model is intentionally given an incorrect function-family label:
\begin{quote}
\small
IMPORTANT: The underlying function is \{wrong\_type\}.
\end{quote}

\subsection{Function families and task instances}
The paper uses several small controlled function families rather than one monolithic benchmark suite.
\begin{itemize}[leftmargin=*]
    \item The static-structure study uses one-dimensional analytic families chosen so that sparse observations leave multiple plausible completions; representative rows in Figure~\ref{fig:static-structure-grid} include linear, quadratic, and sinusoidal examples.
    \item The protocol and uncertainty-alignment studies emphasize Gaussian and logistic families because they provide clear underdetermined regions between and beyond sparse observations.
    \item The sequential-updating study uses one-dimensional tasks with observation orders that can be manipulated by informativeness or conflict.
    \item The downstream optimization study uses the two-dimensional Branin function and a shifted Branin variant.
\end{itemize}
The common design principle is that the observation set is intentionally sparse enough for prompt text and protocol to matter.

\subsection{Output processing and repeated sampling}
For \pw{}, each completion is parsed as a single numeric prediction. For \joint{}, each completion is parsed as a numeric list whose length must match the query list length. Throughout the paper, outputs are requested at the same displayed precision as the examples. This keeps formatting variance from masquerading as predictive variance.

Repeated querying is used for two purposes: to estimate sampling-based dispersion and to construct empirical completion distributions. In the protocol-comparison experiments we use 50 repeated samples per condition, which is the setting used in the figures in the main text. When a completion cannot be parsed into the required numeric format, it is treated as invalid for metrics that require numeric predictions; the same parsing rule is applied uniformly across conditions.

\section{Models in the Figure}
\label{app:models}

Table~\ref{tab:models-in-figure} lists the models used in the paper.

\begin{table}[h]
\centering
\caption{Models shown in the figure.}
\label{tab:models-in-figure}
\begin{tabular}{ll}
\toprule
\textbf{Legend Name} & \textbf{Full Model Name} \\
\midrule
GPT-4o & GPT-4o \\
GPT-4o-mini & GPT-4o mini \\
GPT-5.4 & GPT-5.4 \\
Qwen3-235B-A22B & Qwen3-235B-A22B \\
Llama-3.3-70B & Llama 3.3 70B Instruct \\
Qwen3.5-397B-A17B & Qwen3.5-397B-A17B \\
Llama4-Maverick-17B-128E & Llama 4 Maverick (17Bx128E) \\
\bottomrule
\end{tabular}
\end{table}

\section{Metric Definitions}
\label{app:metrics}

This section defines the metrics used in the main paper. Each metric is a functional of either a single elicited completion curve or an empirical sample from the protocol-conditioned surrogate belief. The purpose of this section is therefore not merely bookkeeping: it clarifies which projection of the elicited surrogate each metric measures.

\subsection{Shape identification and winner rate}
\label{app:shape-id}
To quantify whether a sampled completion recovers the correct \emph{shape family}, we fit a small set of candidate families to the predicted curve and select the best one by corrected Akaike information criterion (AICc).

Let
\[
\mathcal{C}=
\{\texttt{gaussian},\texttt{sinusoidal},\texttt{quadratic},\texttt{linear}\}.
\]
For a trial $t$, the predicted family is
\begin{equation}
\widehat{\mathrm{family}}_t
=
\operatorname*{arg\,min}_{c\in\mathcal{C}} \mathrm{AICc}_{t,c}.
\end{equation}
For candidate family $c$,
\begin{equation}
\mathrm{AIC}_{t,c}
=
 n\log\!\left(\frac{\mathrm{RSS}_{t,c}}{n}+10^{-12}\right)+2K_c,
\end{equation}
\begin{equation}
\mathrm{AICc}_{t,c}
=
\mathrm{AIC}_{t,c}+\frac{2K_c(K_c+1)}{\max(n-K_c-1,1)},
\end{equation}
where $n$ is the number of queried points, $\mathrm{RSS}_{t,c}$ is the residual sum of squares for family $c$, and $K_c=k_c+1$ is the total parameter count, including the residual variance parameter. The mean-function parameter count is
\[
k_c=
\begin{cases}
2, & \text{linear},\\
3, & \text{quadratic},\\
4, & \text{gaussian or sinusoidal}.
\end{cases}
\]
For a group of trials $G$, the winner rate is
\begin{equation}
\mathrm{WinnerRate}(G)
=
\frac{1}{|G|}\sum_{t\in G}
\mathbf{1}[\widehat{\mathrm{family}}_t=\mathrm{family}^{\star}_t].
\end{equation}
Winner rate is a shape-level metric: it rewards recovery of the correct family even when pointwise error is not minimal. In the terminology of the main paper, it is a structural-identification functional on sampled draws from the elicited surrogate belief.

\subsection{NRMSE}
\label{app:nrmse}
For each trial $t$, pointwise fit quality is measured by normalized root mean squared error
\begin{equation}
\mathrm{NRMSE}_t
=
\frac{\sqrt{\frac{1}{m}\sum_{j=1}^{m}(\hat y_t(x_j)-f_t^{\star}(x_j))^2}}
{y_t^{\max}-y_t^{\min}+\varepsilon},
\end{equation}
where $\{x_j\}_{j=1}^{m}$ is the evaluation grid and $y_t^{\max},y_t^{\min}$ are the maximum and minimum ground-truth values on that grid.

\subsubsection{Constraint Satisfaction Rate (CSR)}
\label{app:csr}

CSR measures whether a predicted query value remains compatible with the stated
function family and the two observed anchors. For each case $c$, let
$(x_{1c},y_{1c})$ and $(x_{2c},y_{2c})$ denote the two anchor points, and let
$(x_{cj},\hat y_{cj})$ denote a predicted query point. We define
\begin{equation}
\mathrm{CSR}_{f,g}
=
\frac{1}{\sum_{c\in\mathcal C_g} n_c}
\sum_{c\in\mathcal C_g}
\sum_{j=1}^{n_c}
\mathbf 1\!\left[
\mathcal F_f
\left(
(x_{1c},y_{1c}),
(x_{2c},y_{2c}),
(x_{cj},\hat y_{cj})
\right)
=1
\right],
\label{eq:csr_feasibility}
\end{equation}
where $\mathcal F_f$ is a family-specific feasibility predicate implemented by
analytic or closed-form checks with numerical tolerances. Thus CSR is not
computed by fitting the predicted point to the ground-truth curve, nor by
measuring prediction error. It asks whether the predicted point could belong to
some member of the stated family that is also compatible with the observed
anchors.

For the linear family, feasibility reduces to the tolerance check
\[
\left|
\hat y_{cj}
-
\left(
y_{1c}
+
\frac{y_{2c}-y_{1c}}{x_{2c}-x_{1c}}
(x_{cj}-x_{1c})
\right)
\right|
\le
\varepsilon_{\mathrm{lin}},
\]
with $\varepsilon_{\mathrm{lin}}=10^{-3}$ in our implementation.

For the Gaussian family, feasibility is checked by an analytic log-shape test.
When the three candidate values are positive and the three $x$ locations are
distinct, we compute the unique quadratic interpolant through
$(x_{1c},\log y_{1c})$, $(x_{2c},\log y_{2c})$, and
$(x_{cj},\log \hat y_{cj})$. The triple is considered Gaussian-feasible only if
the resulting quadratic has negative curvature, up to numerical tolerance
$\varepsilon_{\mathrm{gauss}}=10^{-12}$.

For the logistic family, feasibility is checked by a three-point analytic
existence test for the no-offset logistic form. The three values must be
positive and have distinct $x$ locations; we then test whether there exists an
upper asymptote $L$ larger than all three values such that the logit-transformed
points
\[
\log\frac{y}{L-y}
\]
are collinear in $x$, up to tolerance
$\varepsilon_{\mathrm{logistic}}=10^{-3}$.

For the quadratic family, any three finite observations at distinct $x$
locations are feasible because a quadratic interpolant exists. If two candidate
points share the same $x$ location, feasibility requires their $y$ values to
agree up to the same numerical tolerance used for repeated-location checks. For
the exponential family with offset, three non-conflicting observations are
treated analogously as feasible under the three-parameter form
$A\exp(rx)+b$.

In Figure~\ref{fig:protocol-summary}C, the relevant comparison is between the
left group (\pw{}) and the right group (\joint{}). Lower CSR under \joint{}
indicates that autoregressive completion more easily weakens adherence to the
observed anchors. Within the pointwise group, CSR remains comparatively high
across prompt styles, which is why we interpret warning-induced diversity in
\pw{} as structured rather than arbitrary. Operationally, CSR is an
anchor-faithfulness functional: it asks whether a predicted point still belongs
to some completion compatible with the stated family and the observed anchors.

\subsubsection{Pointwise Fit Recoverability Score (FRS)}
\label{app:frs}

FRS measures whether the predicted outputs for one case can still be well explained by a single fitted curve within the target family. For each case \(c\), we first fit a curve within family \(f\) and compute
\[
\mathrm{MSE}_c=\frac{1}{n_c}\sum_{j=1}^{n_c}\left(\hat y_{cj}-h_f(x_{cj};\hat\theta_c)\right)^2,
\qquad
s_c=\exp\!\left(-\frac{\mathrm{MSE}_c}{\tau_f}\right),
\qquad
\mathrm{FRS}_{f,g}=\frac{1}{|\mathcal C_g|}\sum_{c\in\mathcal C_g}s_c.
\]
Here \(\tau_f\) is a family-specific scale parameter that keeps the score comparable across cases.

Panel~B in Figure~\ref{fig:protocol-summary} should be read within the pointwise setting only. Lower FRS means that repeated pointwise completions are less tightly concentrated around a single recoverable family fit. In the main text, we therefore use a reduction in FRS as evidence that task-relevant warnings broaden the candidate completion distribution and make repeated outputs less stable. Operationally, FRS is a one-curve recoverability functional on repeated samples from the elicited belief.

\subsection{Sequential-updating metrics}
For a trial and evidence order $o$, let $\bar c_t$ denote mean token-level confidence over the fixed query grid after step $t$. We report:
\begin{itemize}[leftmargin=*]
    \item stepwise accuracy $\mathrm{Acc}_t=1-\mathrm{NRMSE}_t$;
    \item mean confidence $\bar c_t$;
    \item confidence-valley timing $t_{\mathrm{valley}}=\operatorname*{arg\,min}_{t}\bar c_t$;
    \item confidence-valley depth $\Delta_{\mathrm{valley}}=\bar c_1-\min_t \bar c_t$.
\end{itemize}
The bar plot in Figure~\ref{fig:sequential-updating} summarizes the empirical distribution of $t_{\mathrm{valley}}$ across trials. The term ``confidence valley'' is purely descriptive: it denotes the step of minimum mean model confidence under a reveal order, not a claim about human cognition.

\subsection{Bayesian-optimization metrics}
At BO step $t$, after selecting one or more new points and updating the dataset, we record the incumbent value
\begin{equation}
y_t^{\mathrm{best}} = \max_{(x_i,y_i)\in D_t} y_i
\end{equation}
in the maximization convention used in the paper. Simple regret is then
\begin{equation}
r_t = f^{\star}(x^{\star}) - y_t^{\mathrm{best}},
\end{equation}
where $x^{\star}$ is the global maximizer of the benchmark objective on the search domain. Final regret is $r_T$ at the end of the BO horizon. In addition to regret curves and final-regret distributions, Figure~\ref{fig:branin-bo} reports representative spatial query traces and within-method uncertainty-versus-error scatter plots as diagnostics of search behavior.

\appendix
\section{Model-Dependent Belief Patterns}
\label{app:belief_patterns}

\begin{figure}[H]
    \centering
    \includegraphics[width=0.8\linewidth]{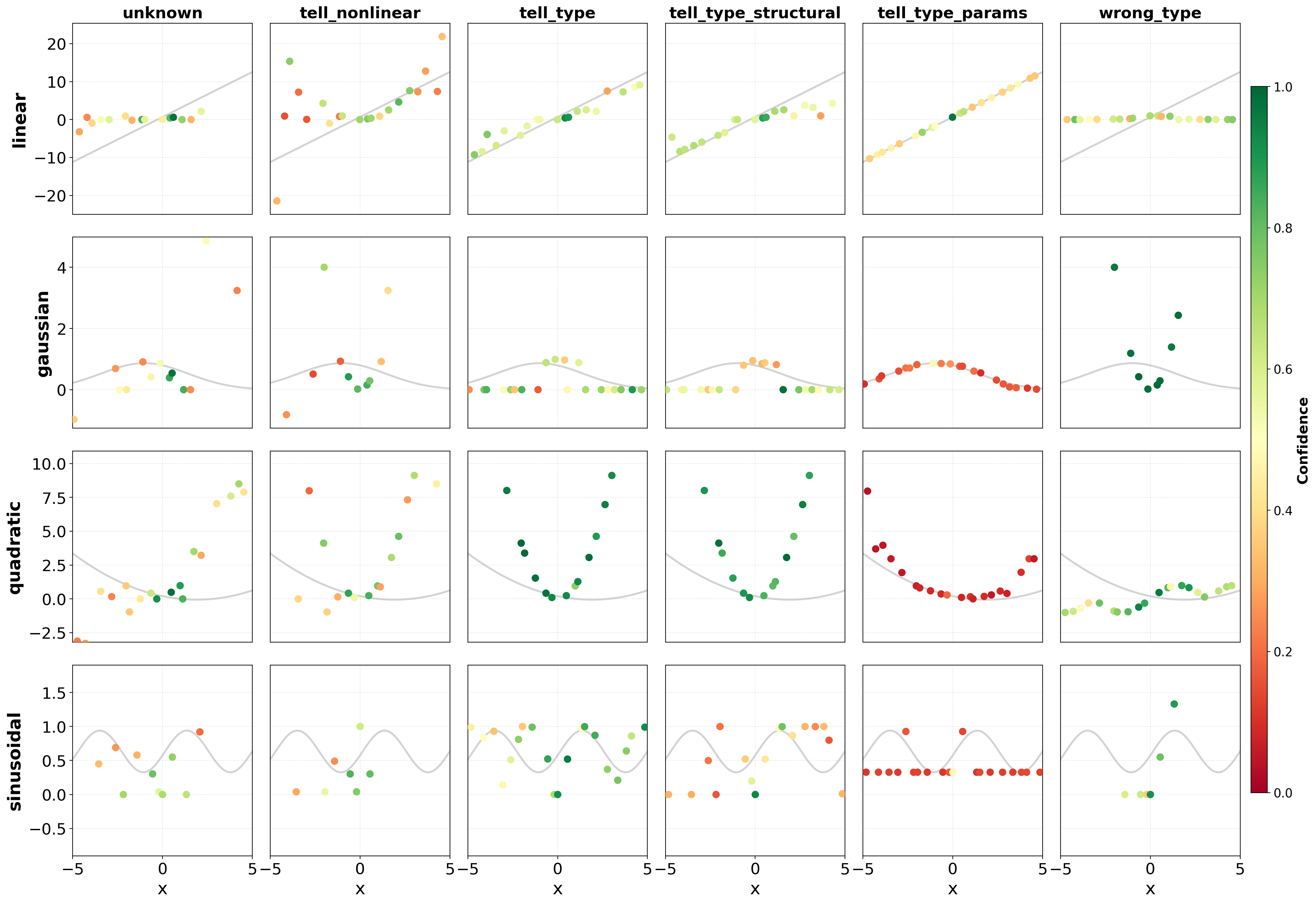}
    \caption{Belief patterns of GPT-4o without observations.}
    \label{fig:gpt4o_belief}
\end{figure}

\begin{figure}[H]
    \centering
    \includegraphics[width=0.8\linewidth]{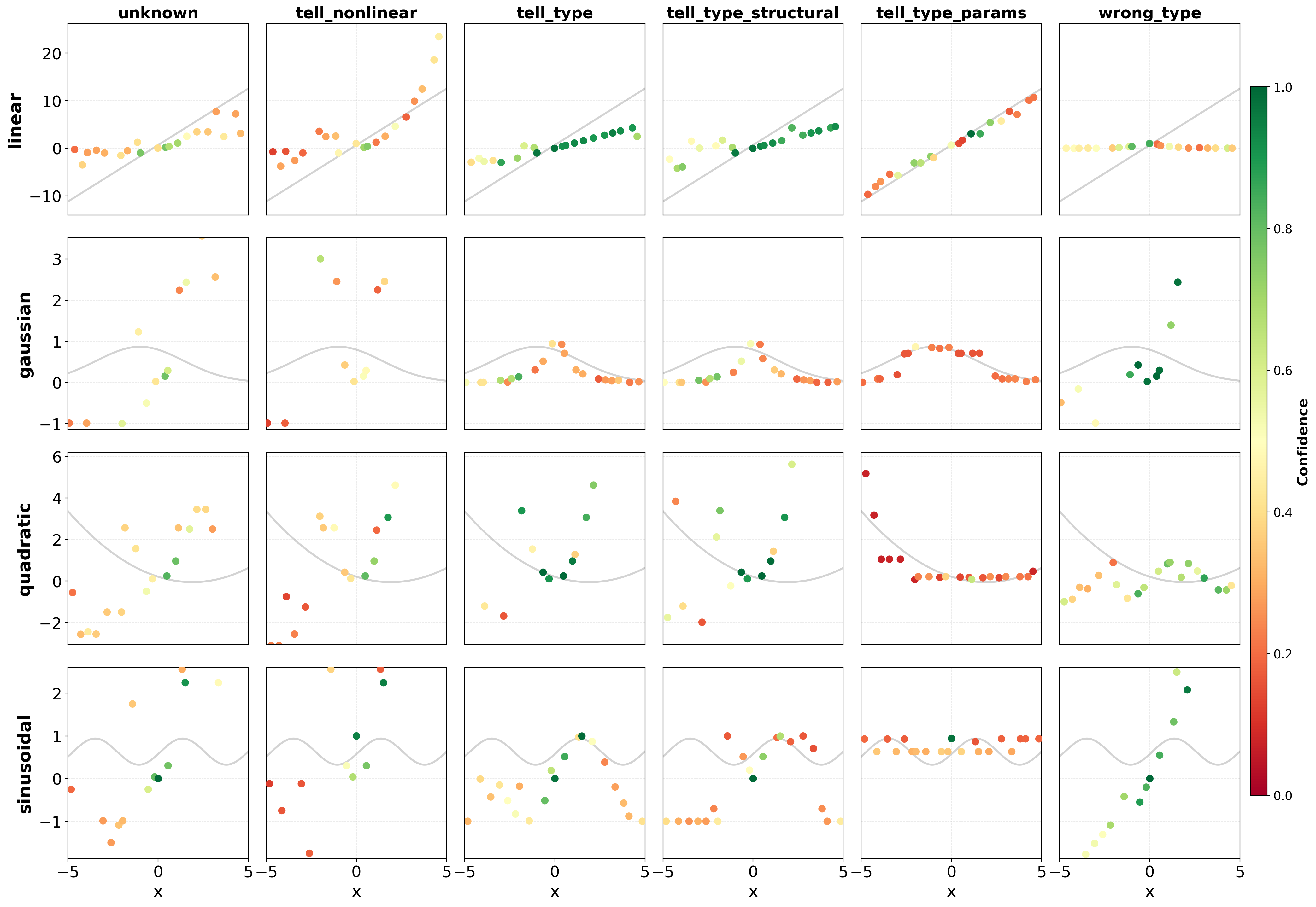}
    \caption{Belief patterns of GPT-4o-mini without observations.}
    \label{fig:gpt4omini_belief}
\end{figure}

Without observations, LLM beliefs may still form structured patterns rather than scattered points. These implicit patterns vary across models, as shown in Figures~\ref{fig:gpt4o_belief} and~\ref{fig:gpt4omini_belief}.

\section{Extrapolation Beyond the Familiar Numerical Range}
\label{app:large-x-extrapolation}

In the main prompt-effect experiments, query inputs are sampled from the interval $[-5,5]$. This range contains small numerical values that are likely to be common in the pretraining distribution of large language models. To examine whether the observed behavior is specific to this familiar numerical range, we ran an additional diagnostic experiment in which the model was asked to extrapolate to much larger input values.

\begin{figure}[H]
    \centering
    \includegraphics[width=0.6\textwidth]{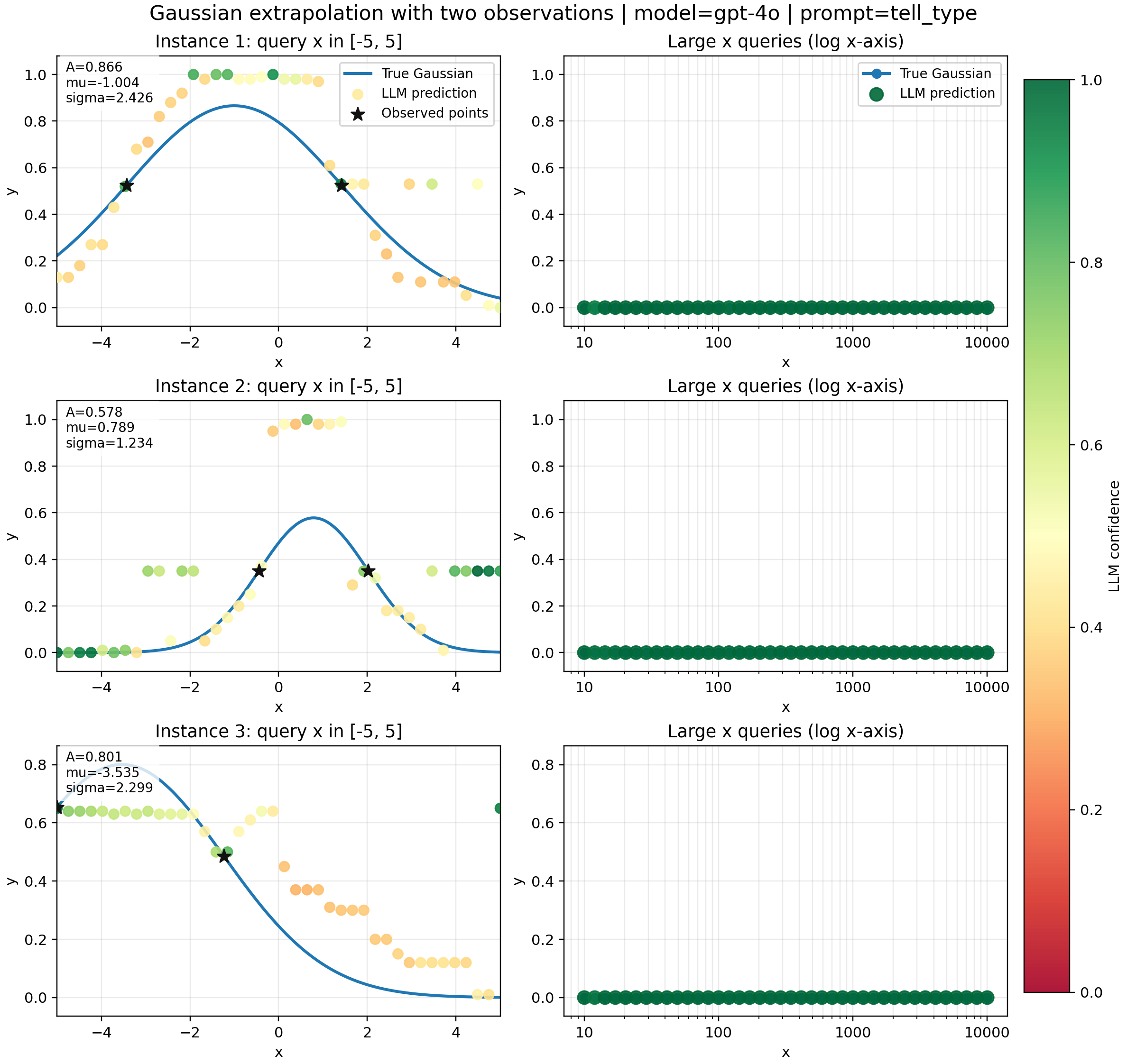}
    \caption{
    Pointwise extrapolation to large input values for Gaussian functions.
    Each row shows one randomly sampled Gaussian instance with two observed points inside $[-5,5]$.
    The left panel shows 40 pointwise queries in $[-5,5]$, and the right panel shows 40 pointwise queries logarithmically spaced in $[10,10^4]$.
    Blue curves denote the true function, black stars denote observed points, and colored markers denote LLM predictions.
    Marker color indicates model confidence, from red (low) to green (high).
    }
    \label{fig:large-x-gaussian}
\end{figure}

\begin{figure}[H]
    \centering
    \includegraphics[width=0.6\textwidth]{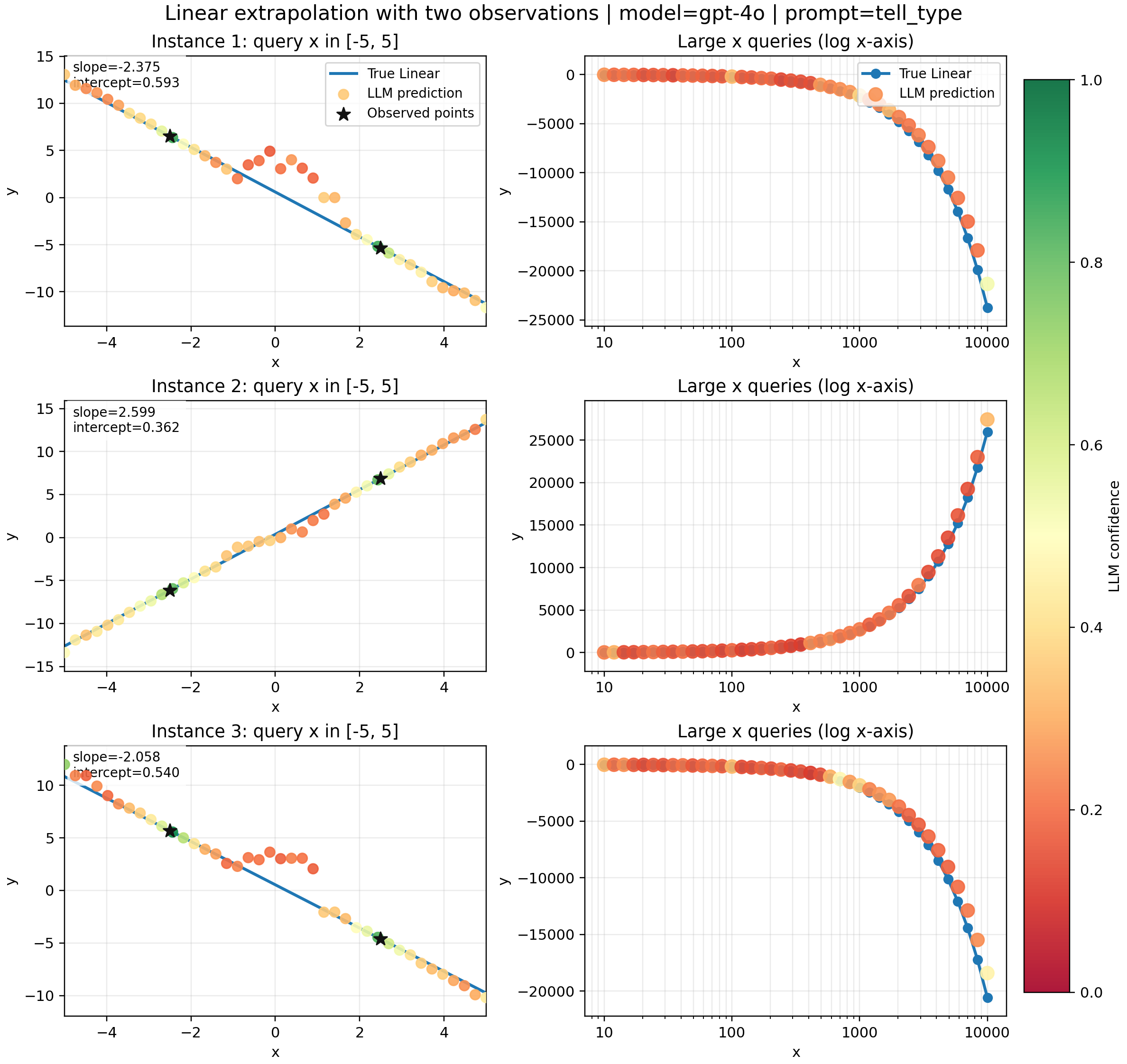}
    \caption{
    Pointwise extrapolation to large input values for linear functions.
    The setup is identical to Figure~\ref{fig:large-x-gaussian}, except that the underlying functions are linear.
    Unlike Gaussian functions, the true linear functions continue to grow or decrease outside the observed range, making this a contrasting case for large-scale extrapolation.
    }
    \label{fig:large-x-linear}
\end{figure}

We considered two one-dimensional function families: Gaussian functions and linear functions. For each family, we sampled three random function instances using the same function generator as in the main experiments. For each instance, the model was given exactly two observed points sampled inside the standard input range $[-5,5]$. We then queried the model pointwise at two sets of locations: 40 evenly spaced points in $[-5,5]$, and 40 logarithmically spaced points in $[10,10^4]$. Each query was issued as a separate prompt, so the model predicted one scalar output value at a time.

The prompt informed the model of the function family and provided the two observed examples. For example, for a Gaussian function, the prompt stated that the underlying function was Gaussian and then asked the model to predict the value at a single query input. This setup isolates pointwise extrapolation behavior while keeping the observation set fixed.

For Gaussian functions, the true function values decay rapidly toward zero outside the range containing the peak. Therefore, a model that correctly extrapolates the Gaussian form should predict values close to zero for large positive inputs. For linear functions, in contrast, the true function continues to grow or decrease linearly outside the observed range. This provides a useful comparison: Gaussian extrapolation requires recognizing saturation or decay, whereas linear extrapolation requires continuing a simple global trend.

Figure~\ref{fig:large-x-gaussian} shows the Gaussian results. Each row corresponds to one sampled Gaussian instance. The left panel shows predictions inside the familiar range $[-5,5]$, while the right panel shows predictions for large query values on a logarithmic $x$-axis. The blue curve shows the true function, black stars indicate the two observed points, and colored markers show the LLM predictions. Marker color denotes the model's confidence, with red indicating low confidence and green indicating high confidence. Across all three instances, the model predicts values close to zero for large positive inputs, consistent with the rapid decay of the Gaussian function.

Figure~\ref{fig:large-x-linear} shows the corresponding results for linear functions. In contrast to the Gaussian case, the correct extrapolation behavior for a linear function is to continue the global linear trend beyond the observed range. The model's predictions in the large-$x$ regime remain broadly consistent with this behavior, suggesting that it does not merely collapse to a generic default response for unfamiliar numerical inputs.

Overall, this diagnostic suggests that the model's pointwise predictions are not restricted to the familiar numerical range $[-5,5]$. Even when queried at much larger values such as $10$--$10^4$, the model often produces estimates that remain consistent with the stated function family. In particular, the Gaussian case tests whether the model extrapolates the bell-shaped function toward zero at large $x$, rather than continuing a local trend observed from the two in-range points. The linear case serves as a contrasting setting where large-magnitude extrapolation is mathematically appropriate.

\section{Verbalizing Shape Bias Under Unknown Function Prompts}
\label{app:verbalization}

The main experiments show that, even when the prompt does not reveal the function family, LLM numerical predictions can exhibit systematic default structure, such as linear or monotonic completions. We therefore ran an auxiliary verbalization experiment to test whether this numerically expressed bias is also available as an explicit verbal judgment.

For each trial, the model was given the same sparse observed points under an unknown-function framing, without being told the true function family or any structural descriptor. Instead of asking for numerical predictions at query locations, we asked the model to describe the pattern in one short sentence, choose one label from a fixed set
\[
\mathcal{Y} =
\left\{
\begin{array}{l}
\texttt{linear\_increasing},\ \texttt{linear\_decreasing},\\
\texttt{bell\_shaped},\ \texttt{quadratic\_upward},\\
\texttt{quadratic\_downward},\ \texttt{periodic},\\
\texttt{unclear}
\end{array}
\right\}.
\]
and report a confidence score between 0 and 1. The prompt therefore tests explicit shape attribution rather than pointwise numerical completion.

\begin{figure*}[h]
    \centering
    \includegraphics[width=0.95\textwidth]{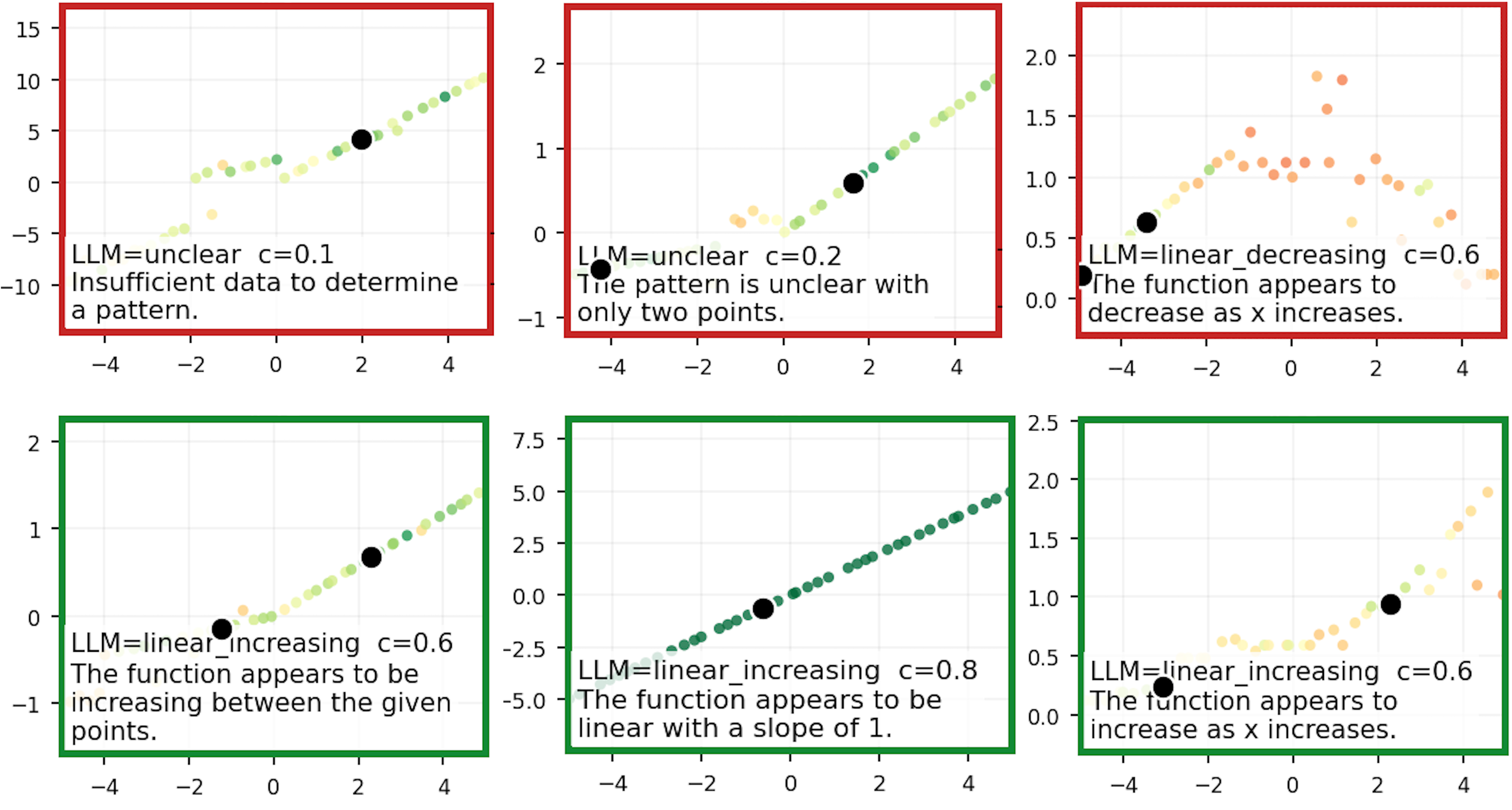}
    \caption{
    \textbf{Verbalized trend labels under unknown-function prompts.}
    The model is given sparse observed points from an unknown one-dimensional function and asked to describe the global pattern, choose a trend label, and report confidence. Black points denote observed examples; colored points show the corresponding numerical pointwise predictions from the unknown-function condition. Red frames indicate verbal labels that are inconsistent with the apparent/generated structure, while green frames indicate agreement. The figure illustrates that numerical completions can exhibit linear or monotonic bias even when the model does not reliably verbalize that bias.
    }
    \label{fig:verbalization-appendix}
\end{figure*}

Figure~\ref{fig:verbalization-appendix} shows representative trials. Black points are the observed in-context examples. The small colored points show the corresponding numerical pointwise predictions from the unknown-function condition, included to make the contrast visible. Text inside each panel reports the model's verbal trend label and confidence. Green frames mark cases where the verbal label agrees with the apparent/generated structure, while red frames mark clear mismatches.

The examples show that explicit verbalization is often more fragile than numerical prediction. In some trials, the numerical predictions form an approximately increasing linear pattern, but the verbal response is \texttt{unclear} with low confidence. In another trial, the visible pattern is increasing, yet the verbal label is \texttt{linear\_decreasing}. Conversely, there are cases where the model correctly labels an increasing linear trend. Thus, the same model can expose a structured numerical completion while failing to state that structure reliably in language.

This supports the distinction used throughout the paper. A pointwise numerical answer can be produced by local interpolation, pattern continuation, or other implicit computations over the context. By contrast, a verbal shape description requires the model to aggregate the observed points into a global, discrete structural hypothesis. The latter is a more explicit reasoning task and is not guaranteed to match the computation that generated the numerical predictions. In this sense, verbal reasoning is not equivalent to the optimizer-facing surrogate belief elicited through numerical queries.

\section{Reference Ambiguity Construction}
\label{app:mcvar}

The reference ambiguity profile $\mcvar(x;D_t)$ is designed to quantify residual ambiguity induced by the observation set within a controlled function family. It is not intended to be the model's true posterior variance.

\subsection{Generic Monte Carlo construction}

Suppose the reference family is parameterized by $\phi\in\Phi$ with function $f_{\phi}$. We begin with a broad proposal distribution $p_0(\phi)$ over plausible parameters and generate Monte Carlo proposals
\begin{equation}
\phi^{(m)} \sim p_0(\phi),
\qquad m=1,\dots,M.
\label{eq:mc_proposals}
\end{equation}
A proposal is retained if it remains consistent with the observations up to tolerance $\varepsilon_{\mathrm{cons}}$, that is, if
\begin{equation}
\max_{(x_i,y_i)\in D_t}|f_{\phi^{(m)}}(x_i)-y_i|
\leq
\varepsilon_{\mathrm{cons}}.
\label{eq:mc_accept}
\end{equation}
Let $\{\phi^{(1)},\ldots,\phi^{(K)}\}$ denote the retained sample-consistent proposals. These retained samples induce an empirical Monte Carlo reference distribution
\begin{equation}
\hat q_D^{\mathrm{MC}}
=
\frac{1}{K}\sum_{k=1}^{K}\delta_{\phi^{(k)}}.
\label{eq:qD_mc}
\end{equation}
The reference ambiguity profile is then defined as
\begin{equation}
\mcvar(x;D_t)
=
\operatorname{Var}_{\phi\sim \hat q_D^{\mathrm{MC}}}[f_{\phi}(x)].
\label{eq:vh_appendix}
\end{equation}
Operationally, this is the empirical variance of the retained function values at each query location:
\begin{equation}
\mcvar(x;D_t)
=
\operatorname{Var}\!\left(
f_{\phi^{(1)}}(x),\ldots,f_{\phi^{(K)}}(x)
\right).
\label{eq:vh_empirical}
\end{equation}

This is a rejection-style Monte Carlo construction over a controlled reference family. Its purpose is to provide an operational signal of residual sample-consistent ambiguity, not to estimate the model's true posterior variance or to define a universally correct uncertainty target.

\subsection{Why a variance profile?}

The sample-consistent family can be summarized in several ways, including pointwise width, envelope area, or posterior-like variance. We use a variance profile for two practical reasons. First, it produces a smooth pointwise ambiguity signal on a fixed query grid, which makes comparison to model-side uncertainty proxies straightforward. Second, the alignment metric in the main text is rank-based, so the precise scale of $\mcvar(x;D_t)$ is less important than whether it ranks ambiguous regions above less ambiguous ones.

For intuition, a related idealized quantity is the width functional
\begin{equation}
W_{\mathcal{H}}(x;D_t)
=
\sup_{f,g\in\mathcal{H}(D_t;\mathcal{F})}|f(x)-g(x)|.
\label{eq:width_appendix}
\end{equation}
Both $W_{\mathcal{H}}$ and $\mcvar$ summarize residual ambiguity in the sample-consistent family, but $\mcvar$ is especially convenient for Monte Carlo approximation and for direct comparison to model-side uncertainty proxies.

\subsection{Families used in the main alignment plots}

In the main uncertainty-alignment experiments we use controlled Gaussian and logistic families. These are useful because sparse observations leave visually interpretable regions of residual ambiguity.

For sparse Gaussian examples, ambiguity is often concentrated around unresolved center and width parameters, so retained sample-consistent functions disagree most strongly near the uncertain peak region. For sparse logistic examples, ambiguity is often concentrated around the transition region, where different retained parameter settings induce different slopes and horizontal shifts.

In both cases, the Monte Carlo construction above yields a pointwise profile $\mcvar(x;D_t)$ that highlights where sample-consistent hypotheses disagree most strongly. This is the reference signal used in Section~\ref{sec:alignment} when evaluating whether model-side uncertainty ranks ambiguous regions above less ambiguous ones.

\section{Decoding and Protocol-Comparison Details}
\label{app:decoding}

\subsection{Observed-point formatting}
All observed examples are rendered with fixed decimal precision, typically two decimals in the prompts shown in the paper. Query outputs are required to follow the same precision. This design reduces the chance that formatting variability is misread as epistemic uncertainty.

\subsection{Parsing rules}
For pointwise prompts, a valid completion is a single scalar numeric value. For joint prompts, a valid completion is a list of numeric values in the requested order. If a completion contains extra text, malformed JSON, or the wrong number of outputs, it is marked invalid for numeric analyses. The same parser is used across prompt styles so that any difference in invalidity rate is itself part of the protocol effect rather than an evaluation artifact.

\subsection{Repeated sampling}
Repeated sampling serves two roles: estimating $u_{\theta}^{\mathrm{sam}}(x)$ and constructing empirical completion distributions for OT analysis. In the protocol-comparison experiments we use 50 repeated samples per condition at the same prompt and temperature configuration. Joint samples yield one vector-valued curve per repetition; pointwise samples are assembled by querying each location independently under the same condition.

\subsection{Why protocol affects uncertainty}
The protocol difference is not merely about formatting. In \joint{} mode, later locations condition on earlier generated outputs, so instability can propagate even if token-level confidence remains high within a sampled sequence. In \pw{} mode, each location is asked independently against the same evidence set, which more cleanly exposes location-wise ambiguity. 

\section{Sequential Updating Details}
\label{app:sequential-details}
\label{app:update-extra}

This appendix describes how the sequential evidence conditions in Section~\ref{sec:update} are constructed, how the GP control is computed, and which auxiliary statistics are used in the sequence-order analysis.

\subsection{Experimental setup}

For each sampled 1D function instance $f^{\star}$, we evaluate the surrogate on a fixed query grid
\begin{equation}
\mathcal{X}_{\mathrm{qry}}=\{x^{(1)},\dots,x^{(41)}\},
\end{equation}
which in the plotted experiment is the 41-point grid over $[-5,5]$. Sequential evidence is then revealed one observation at a time for $T$ steps. For Figure~\ref{fig:sequential-updating}, we use $T=20$ observations, six 1D function families (\texttt{gaussian}, \texttt{logistic}, \texttt{exponential}, \texttt{quadratic}, \texttt{linear}, and \texttt{sinusoidal}), and 10 random instances per family and condition, yielding 300 total sequential runs, or 60 runs per condition. The plotted LLM results use pointwise querying: each query $x\in\mathcal{X}_{\mathrm{qry}}$ is asked separately and token-level confidence is extracted from the numeric response.

\subsection{Candidate pool and evidence-order conditions}

For each function instance, we first construct a candidate pool of $T$ observations by evaluating the ground-truth function on a uniform grid over the function domain,
\begin{equation}
\mathcal{C}=\{(x_i,y_i)\}_{i=1}^{T},
\qquad y_i=f^{\star}(x_i).
\end{equation}
All non-conflict conditions use exactly this same pool. Therefore, any difference between \texttt{high\_info\_first}, \texttt{low\_info\_first}, and \texttt{random} is caused by evidence order rather than by observing different points.

The main paper reports five evidence-order conditions:
\begin{itemize}
\item \texttt{high\_info\_first}: observations judged most informative are revealed early;
\item \texttt{low\_info\_first}: observations judged least informative are revealed early;
\item \texttt{random}: a random permutation of the same candidate pool;
\item \texttt{conflict\_middle}: a contradictory observation is inserted mid-sequence;
\item \texttt{conflict\_late}: a contradictory observation is inserted near the end.
\end{itemize}

The summary plots in Figure~\ref{fig:sequential-updating} aggregate 60 runs per condition, as indicated in the figure labels.

\subsection{Operationalizing informativeness}

To rank candidate observations, we use a fixed-hyperparameter GP reference model and score how much a candidate location would reduce posterior spread on the query grid. Let $D$ denote the current revealed prefix. For a candidate location $x$, we compute a global uncertainty-reduction term
\begin{equation}
g_{\mathrm{global}}(x;D)
=
\frac{1}{|\mathcal{X}_{\mathrm{qry}}|}
\sum_{x'\in\mathcal{X}_{\mathrm{qry}}}
\left[\sigma_D(x')-\sigma_{D\cup\{(x,0)\}}(x')\right],
\end{equation}
and a local term centered at $x$,
\begin{equation}
g_{\mathrm{local}}(x;D)
=
\frac{1}{|\mathcal{X}_{\mathrm{qry}}|}
\sum_{x'\in\mathcal{X}_{\mathrm{qry}}}
w(x',x)
\left[\sigma_D(x')-\sigma_{D\cup\{(x,0)\}}(x')\right],
\end{equation}
where $w(x',x)$ is a Gaussian weight centered at $x$. The final information score is
\begin{equation}
g(x;D)=0.35\,g_{\mathrm{global}}(x;D)+0.65\,g_{\mathrm{local}}(x;D).
\label{eq:info-score}
\end{equation}
We use a pseudo-target value $0$ in this ranking step because, under fixed kernel hyperparameters, the posterior variance reduction depends on the candidate location rather than the realized target value.

This score operationalizes informativeness as ambiguity reduction on the evaluation grid: points are informative when they collapse posterior spread broadly or in a locally consequential region. The non-conflict orders are then defined by greedy selection under this score:
\begin{itemize}
\item \texttt{high\_info\_first}: greedily reveal the remaining candidate with largest $g(x;D)$;
\item \texttt{low\_info\_first}: greedily reveal the remaining candidate with smallest $g(x;D)$;
\item \texttt{random}: reveal a random permutation of the same candidate pool.
\end{itemize}
Operationally, \texttt{high\_info\_first} tends to place globally constraining points early, whereas \texttt{low\_info\_first} delays those points and initially reveals more redundant or locally limited support. This is why the GP control improves fastest in \texttt{high\_info\_first}, slowest in \texttt{low\_info\_first}, and why the LLM is forced to revise earlier in the former condition.

\subsection{Conflict observations}

The conflict conditions are designed to test belief revision rather than coverage alone. They are built from a supportive high-information sequence of length $T-1$. Let the intended conflict insertion step be
\begin{equation}
t_c=
\begin{cases}
\lfloor T/2 \rfloor + 1, & \texttt{conflict\_middle},\\
T-1, & \texttt{conflict\_late}.
\end{cases}
\end{equation}
Using only the supportive prefix before step $t_c$, we fit the GP reference model and choose the conflict location at the point where the current surrogate is most confident,
\begin{equation}
x_c=\operatorname*{arg\,min}_{x\in\mathcal{X}_{\mathrm{qry}}}\sigma^{\mathrm{GP}}_{t_c-1}(x).
\end{equation}
Equivalently, this is the point where a contradictory observation will be maximally surprising relative to the current prefix.

Let
\begin{equation}
\mu_c=\mu^{\mathrm{GP}}_{t_c-1}(x_c),
\qquad
y_c^{\star}=f^{\star}(x_c),
\qquad
s_{\mathrm{obs}}=\max\!\bigl(\mathrm{Std}(y_{1:t_c-1}),\,0.2\bigr).
\end{equation}
The magnitude of the contradiction is then chosen by
\begin{equation}
\Delta_c
=
\lambda\Bigl(
0.8\,s_{\mathrm{obs}}
+0.6\,\sigma^{\mathrm{GP}}_{t_c-1}(x_c)
+0.7\,|\mu_c-y_c^{\star}|
+0.25
\Bigr),
\end{equation}
where $\lambda$ is a scale parameter (\texttt{noise\_scale} in code; $\lambda=3.0$ in the plotted experiment), together with the lower bound
\begin{equation}
\Delta_{\min}=\max(0.8,\,0.35\lambda).
\end{equation}
Finally, the contradictory target is defined as
\begin{equation}
y_c=\mu_c+\eta\,\max(\Delta_c,\Delta_{\min}),
\qquad
\eta\in\{-1,+1\}.
\end{equation}
Thus, the conflict point is not merely a low-information point revealed late; it is an explicitly inconsistent observation chosen to overturn the current surrogate. This is why the conflict conditions produce abrupt accuracy drops rather than the slower revision seen in \texttt{low\_info\_first}.

\subsection{Per-step LLM evaluation}

At each step $t$, the current prefix $D_t^{(o)}$ is presented to the LLM and the model is queried on the fixed grid $\mathcal{X}_{\mathrm{qry}}$. Mean token-level confidence is
\begin{equation}
\bar c_t
=
\frac{1}{|\mathcal{X}_{\mathrm{qry}}|}
\sum_{x\in\mathcal{X}_{\mathrm{qry}}}
c_{\theta}^{\mathrm{tok}}(x;D_t^{(o)},p,\pi).
\end{equation}
Predictive fit is evaluated on the same grid using
\begin{equation}
\mathrm{Acc}_t = 1-\mathrm{NRMSE}_t,
\end{equation}
where $\mathrm{NRMSE}_t$ is normalized by the range of the true function values on the query grid. We also summarize the timing of the confidence minimum by
\begin{equation}
t_{\mathrm{valley}}
=
\operatorname*{arg\,min}_{1\le t\le T}\bar c_t.
\end{equation}

\subsection{GP control replay}

To separate model-specific updating from the geometry of the observation sequence, we replay exactly the same prefixes with a GP control using fixed kernel hyperparameters. This control provides a geometric baseline for how sequential evidence alone should affect fit and uncertainty.

For the GP replay, we compute
\begin{equation}
\mathrm{Acc}^{\mathrm{GP}}_t = 1-\mathrm{NRMSE}^{\mathrm{GP}}_t
\end{equation}
on the same query grid. Here $\mathrm{NRMSE}^{\mathrm{GP}}_t$ is normalized by the range of the true function values on $\mathcal{X}_{\mathrm{qry}}$, as in the LLM accuracy calculation.

We also report a GP uncertainty proxy,
\begin{equation}
\bar c^{\mathrm{GP}}_t
=
\operatorname{clip}_{[0,1]}
\left(
1-\frac{1}{|\mathcal{X}_{\mathrm{qry}}|}
\sum_{x\in\mathcal{X}_{\mathrm{qry}}}\sigma_t^{\mathrm{GP}}(x)
\right).
\end{equation}
In this replay, the GP uses fixed kernel hyperparameters with unit signal variance, fixed length scale, no hyperparameter optimization, and no target standardization. Thus $\sigma_t^{\mathrm{GP}}(x)$ is interpreted on the fixed unit-prior GP scale, rather than as an uncertainty measured in the raw target-function units. The clipping only ensures that the displayed proxy lies on the same bounded scale as confidence; it is not used to fit or update the GP.

The dashed curves in Figure~\ref{fig:sequential-updating} correspond to this replay. If a pattern appears in both the LLM and GP curves, it is largely explained by the geometry of the evidence order itself. If the LLM exhibits an additional dip, lag, or mismatch between confidence and accuracy beyond the GP baseline, that extra effect reflects model-side belief updating rather than feasible-set geometry alone.

\subsection{Auxiliary stepwise statistics}

In addition to confidence and accuracy trajectories, we record two GP-based stepwise reference quantities. First, the surprise of the newly revealed observation under the previous GP belief,
\begin{equation}
\mathrm{Conflict}_t
=
-\log p_{\mathrm{ref}}(y_t \mid x_t, D_{t-1}),
\end{equation}
which is large when the new point is inconsistent with the previously supported trend. Second, we record a GP-based information-gain statistic,
\begin{equation}
\mathrm{InfoGain}_t
=
\frac{1}{|\mathcal{X}_{\mathrm{qry}}|}
\sum_{x\in\mathcal{X}_{\mathrm{qry}}}
\bigl[
L_t(x)-L_{t-1}(x)
\bigr],
\end{equation}
where
\begin{equation}
L_t(x)=\log \max_y p_t^{\mathrm{ref}}(y\mid x).
\end{equation}
These quantities are used only for analysis and interpretation; the main figure focuses on confidence and accuracy trajectories.

\subsection{Trajectory classes}

For the supplementary order-comparison panels, we also summarize three coarse trajectory classes: monotone-up (M), U-shape (U), and inverted-U-then-up (I). These labels are descriptive only. They are used to summarize the prevalence of early confidence dips and recoveries across repeated runs, but they are not used as optimization objectives, supervision signals, or training labels.

\section{Bayesian-Optimization Setup Details}
\label{app:details}

This appendix gives the exact experimental settings for the downstream BO studies in Section~\ref{sec:bo}. The purpose of these experiments is not broad benchmarking, but to test whether protocol-induced differences in surrogate belief lead to different acquisition decisions and optimization outcomes.

\subsection{Branin consequence study}

\paragraph{Methods.}
We compare five methods: GP, Random, LLM (Pointwise Neutral), LLM (Pointwise Under\-determination-aware), and LLM (Joint Neutral). In all LLM conditions, the model is not told the underlying function family; the target is always presented as an unknown function. The LLM used in this section is GPT-4o.

\paragraph{Representative LLM prompts.}
The Branin and shifted-Branin experiments use the same prompt format; only the observed coordinates and values differ from run to run. For readability, we show compact templates rather than full run-specific prompts.

\paragraph{\pw{} \neutral{}.}
\begin{quote}
\small
You are a function approximator. Given examples of \(((x_1,x_2),y)\) pairs from an unknown 2D function, predict the \(y\) value for a new \((x_1,x_2)\) input.\\
\\
Rules:\\
1. Output only the numeric \(y\) value.\\
2. Use the same precision as the examples.\\
3. Do not include explanation or extra text.\\
\\
Here are example points from an unknown 2D function:\\
\texttt{[1]} \(x=(x_{1}^{(1)},x_{2}^{(1)})\) \(y=y^{(1)}\)\\
\texttt{[2]} \(x=(x_{1}^{(2)},x_{2}^{(2)})\) \(y=y^{(2)}\)\\
\ldots\\
\\
Given the pattern, predict the \(y\) value for:\\
\(x=(x_{1}^{\mathrm{query}},x_{2}^{\mathrm{query}})\), \(y=\)
\end{quote}

\paragraph{\pw{} Under\-determination-aware.}
\begin{quote}
\small
Here are example points from an unknown 2D function.\\
WARNING: This problem is strongly underdetermined. The observed points do not uniquely determine the underlying surface.\\
Multiple substantially different surfaces, with different slopes, curvature, and local extrema, can fit these same observations equally well.\\
Do not assume a single confident interpretation from sparse evidence.\\
Treat this as a high-uncertainty inference task and make a cautious prediction that reflects ambiguity rather than overconfident extrapolation.\\
\\
\texttt{[1]} \(x=(x_{1}^{(1)},x_{2}^{(1)})\) \(y=y^{(1)}\)\\
\texttt{[2]} \(x=(x_{1}^{(2)},x_{2}^{(2)})\) \(y=y^{(2)}\)\\
\ldots\\
\\
Given this uncertainty, predict the \(y\) value for:\\
\(x=(x_{1}^{\mathrm{query}},x_{2}^{\mathrm{query}})\), \(y=\)
\end{quote}

\paragraph{\joint{} \neutral{}.}
\begin{quote}
\small
You are a function approximator. Given examples of \(((x_1,x_2),y)\) pairs from an unknown 2D function, predict the \(y\) values for a list of new \((x_1,x_2)\) inputs.\\
Return strict JSON only. Output must contain exactly one key ``y'' with a list of numeric values in query order.\\
\\
Here are example points from an unknown 2D function:\\
\texttt{[1]} \(x=(x_{1}^{(1)},x_{2}^{(1)})\) \(y=y^{(1)}\)\\
\texttt{[2]} \(x=(x_{1}^{(2)},x_{2}^{(2)})\) \(y=y^{(2)}\)\\
\ldots\\
\\
Predict \(y\) for each query in order.\\
Query list:\\
\texttt{[1]} \(x=(u_{1}^{(1)},u_{2}^{(1)})\)\\
\texttt{[2]} \(x=(u_{1}^{(2)},u_{2}^{(2)})\)\\
\ldots\\
\\
Return: \texttt{\{"y":[y1,\ldots,yk]\}}
\end{quote}

\paragraph{Acquisition rule.}
At each BO step $t$, each method scores candidates and selects the top-ranked points. For the GP baseline, we use the standard UCB score
\[
a_t^{\mathrm{GP}}(x)=\mu_t^{\mathrm{GP}}(x)+\beta\sigma_t^{\mathrm{GP}}(x),
\]
where $\sigma_t^{\mathrm{GP}}(x)$ is the GP posterior predictive standard deviation and $\beta=3.0$.

For the LLM conditions, we use a UCB-style ranking score
\[
a_t^{\mathrm{LLM}}(x)=\mu_t^{\mathrm{LLM}}(x)+\lambda n_t(x),
\]
where $\mu_t^{\mathrm{LLM}}(x)$ is the numeric prediction and $n_t(x)$ is the mean token negative log-probability of the numeric output. We treat $n_t(x)$ as a dimensionless exploration feature, not as an objective-scale posterior standard deviation. The coefficient $\lambda$ is fixed across LLM conditions within the task, so the comparison isolates prompt/protocol effects under the same decision rule.

\paragraph{Budget and candidate sets.}
Each trial begins with $5$ initial observations and then runs for $10$ BO steps. At every step, a fresh candidate pool of $80$ points is sampled uniformly from the search domain, and the method selects the top $3$ points under the acquisition score. These selected points are evaluated and added to the observation set. All methods are evaluated on the same initial observations and the same per-step candidate pools within each trial.

\paragraph{Objectives.}
We consider two related 2D objectives on the domain $[-5,10]\times[0,15]$.

The first is the canonical Branin function, a standard benchmark for Bayesian optimization.

The second is a shifted Branin variant constructed to preserve substantial local resemblance to canonical Branin while moving the optimum to a different location. The purpose of this design is to test whether the LLM optimizer relies on a strong prior over the familiar Branin landscape: if so, such a prior should help on canonical Branin but become misleading once the optimum is displaced.

\paragraph{Construction of modified Branin.}
The shifted-Branin objective is constructed by evaluating the canonical Branin function after smooth monotone warps of the two input coordinates. Let
\[
u=\frac{x_1+5}{15}, \qquad v=\frac{x_2}{15},
\]
so that $(u,v)\in[0,1]^2$. The warped coordinates are
\[
\tilde x_1=-5+15\,\phi_1(u), \qquad \tilde x_2=15\,\phi_2(v),
\]
where each warp is an endpoint-normalized logistic transform
\[
\phi(z;k,b)
=
\frac{\sigma(k(z-b))-\sigma(-kb)}
{\sigma(k(1-b))-\sigma(-kb)},
\qquad
\sigma(t)=\frac{1}{1+e^{-t}}.
\]
In the implementation, the two axes use different parameters:
\[
\phi_1(u)=\phi(u;7.5,0.33), \qquad
\phi_2(v)=\phi(v;5.5,0.62).
\]
The shifted objective is then
\[
f_{\mathrm{shift}}(x_1,x_2)=f_{\mathrm{Branin}}(\tilde x_1,\tilde x_2).
\]
Thus the function preserves the overall Branin topology, but the basin locations and shapes are displaced by anisotropic nonlinear warping.

\paragraph{Camouflaged initialization for modified Branin.}
In the shifted-Branin condition, the objective is constructed as a Branin-like landscape with an additional hidden well, so that its global optimum is displaced from the canonical Branin minima. Writing the canonical Branin objective as $f_{\mathrm{B}}(x)$ and the shifted objective as $f_{\mathrm{shift}}(x)$, the initialization set is not sampled arbitrarily. Instead, candidate initial points are drawn uniformly from the domain and retained only if
\[
|f_{\mathrm{shift}}(x)-f_{\mathrm{B}}(x)| < \tau,
\]
with a strict threshold $\tau=2$ and a relaxed threshold $\tau=2.5$ if too few points satisfy the stricter condition. Thus, each initial observation is individually chosen so that the shifted objective remains close in value to canonical Branin.

Among the retained candidates, the final set of initial points is chosen to minimize a camouflage score that favors: (i) small average value gap between shifted and canonical Branin, (ii) low pairwise ranking disagreement between the two objectives on the selected points, and (iii) low proximity to the hidden well, while also encouraging spatial spread. Formally, for an initialization set $S$,
\[
\mathrm{Score}(S)
=
\frac{1}{|S|}\sum_{x\in S}|f_{\mathrm{shift}}(x)-f_{\mathrm{B}}(x)|
+2\,\mathrm{RankMismatch}(S)
+\mathrm{WellProximity}(S)
-0.3\,\mathrm{Spread}(S),
\]
and the set with the lowest score is used. This creates a deliberately camouflaged prefix of evidence: the early observations are broadly consistent with the standard Branin landscape even though the true optimum lies elsewhere. The goal is to test whether the optimizer becomes anchored to the familiar prior and is therefore drawn toward a decoy basin.

\paragraph{Metrics.}
Our main metric is best-so-far simple regret after each BO step. We write the objective in reward form, so larger values are better and the optimal value is normalized to $0$. Regret is therefore
\[
r_t = y^\star - \max_{s \le t} y_s,
\]
with $y^\star = 0$. We also record whether the selected batch contains a top-$10\%$ candidate in the current pool and compute the Spearman rank correlation between surrogate mean predictions and true candidate values, although these auxiliary diagnostics are not the main focus of Section~\ref{sec:bo}.

\paragraph{Repeated trials.}
Each condition is repeated over $10$ trials. Reported curves show the evolution of best-so-far regret across BO steps, and the figure also compares the distribution of final regret across trials.

\subsection{XGBoost HPO}

The realistic HPO experiment is a fixed-library Bayesian optimization task over XGBoost hyperparameters. The main-text panel reports the California Housing regression task only. The hidden objective is XGBoost validation performance on a fixed train/validation split. For regression, the optimizer stores negative validation RMSE, so larger values are better.

\paragraph{Search space and library.}
The search space contains ten XGBoost hyperparameters: learning rate, maximum depth, minimum child weight, row subsampling, column subsampling, L2 regularization, L1 regularization, split penalty, number of boosting rounds, and histogram bin count. The ranges are deliberately wide, including underfitting and over-regularized regions, so that random search is not trivially competitive. A Sobol library of $128$ configurations is precomputed and reused by all methods. BO seeds change the initial observations and search trajectory, not the underlying objective.

\paragraph{Budget and candidate sets.}
Each run starts from $3$ initial observations and then performs $20$ one-point BO steps. The per-step candidate pool is shared across methods within a seed. In the plotted run, the candidate-pool size is $128$, which means that after removing already observed configurations, each method scores essentially all remaining configurations in the fixed library.

\paragraph{Methods and acquisition.}
The plotted methods are Random, SKOpt-style GP+EI, Optuna-style TPE, SMAC3-style random forest+EI, LLM Pointwise Neutral, and LLM Joint Neutral. The LLM model is Qwen3.5-397B-A17B. The saved run also contains a pointwise warning condition, but the main panel focuses on the neutral protocol comparison. The classical surrogate baselines use their standard expected-improvement implementations. For LLM methods, the predicted mean is the model's predicted validation objective, and the mean negative log-probability of the numeric output tokens is used as a heuristic exploration feature in an EI-style ranking score. This LLM score is used only to rank candidates within the same BO step; it should not be interpreted as calibrated expected improvement under a Gaussian posterior.

\paragraph{LLM-visible information.}
The LLM sees the model family (XGBoost), task type, sample count, feature count, observed hyperparameter configurations with their objective values, and either one candidate (\pw{}) or a list of candidates (\joint{}). It does not see the dataset name, OpenML identifier, or any text identifying California Housing.

\paragraph{Representative LLM prompts.}
For readability, we show compact templates rather than full candidate lists.

\paragraph{\pw{} \neutral{}.}
\begin{quote}
\small
You are optimizing XGBoost hyperparameters from past trials.\\
Predict the validation objective value for one candidate configuration.\\
Objective: RMSE (lower is better).\\
\\
Model family: XGBoost gradient-boosted trees\\
Task type: regression\\
Samples: [number of training examples]\\
Features: [number of input features]\\
\\
Observed trials:\\
\texttt{[1]} [hyperparameter configuration 1] \(\rightarrow\) objective (RMSE) = [observed value 1]\\
\texttt{[2]} [hyperparameter configuration 2] \(\rightarrow\) objective (RMSE) = [observed value 2]\\
\ldots\\
\\
Candidate:\\
\emph{candidate hyperparameter configuration}\\
\\
Return only one numeric objective value.
\end{quote}

\paragraph{\joint{} \neutral{}.}
\begin{quote}
\small
You are optimizing XGBoost hyperparameters from past trials.\\
Predict validation objective values for multiple candidate configurations.\\
Objective: RMSE (lower is better).\\
\\
Model family: XGBoost gradient-boosted trees\\
Task type: regression\\
Samples: [number of training examples]\\
Features: [number of input features]\\
\\
Observed trials:\\
\texttt{[1]} [hyperparameter configuration 1] \(\rightarrow\) objective (RMSE) = [observed value 1]\\
\texttt{[2]} [hyperparameter configuration 2] \(\rightarrow\) objective (RMSE) = [observed value 2]\\
\ldots\\
\\
Candidates in order:\\
\texttt{[1]} [candidate configuration 1]\\
\texttt{[2]} [candidate configuration 2]\\
\ldots\\
\\
Return strict JSON only: \texttt{\{"scores":[s1,s2,s3,\ldots]\}}
\end{quote}

\paragraph{Outcome summary.}
Figure~\ref{fig:hpo} reports raw best-so-far simple regret relative to the best configuration in the fixed library. Mean final raw regret over ten seeds is $0.0043$ for SMAC3, $0.0087$ for GP+EI, $0.0107$ for LLM Pointwise Neutral, $0.0108$ for LLM Joint Neutral, $0.0141$ for Random, and $0.0217$ for TPE. Thus the LLM pointwise surrogate is competitive with standard BO baselines on this task, but SMAC3 is more reliable by the end of the budget.

\subsection{Battery cathode-design fixed-pool BO}
\label{sec:battery-pybamm}

The battery experiment is a fixed-pool BO task over PyBaMM DFN simulations of a redesigned Chen2020 full cell. We optimize two active cathode-design variables,
\[
L_{\mathrm{pos}}\in[4.0\times10^{-5},1.5\times10^{-4}]~\mathrm{m},
\qquad
\epsilon_{\mathrm{pos}}\in[0.22,0.45],
\]

\paragraph{Objective.}
For each candidate design, PyBaMM evaluates the delivered volumetric energy density at both low-rate and high-rate discharge, denoted by
\(\mathrm{ED}(0.5\mathrm{C})\) and \(\mathrm{ED}(3\mathrm{C})\), in Wh/L. We normalize these two quantities by rate-specific reference values,
\[
Q_{\mathrm{low}} = 943.05~\mathrm{Wh/L},
\qquad
Q_{\mathrm{high}} = 509.35~\mathrm{Wh/L},
\]
defined as the 90th percentiles of successful pilot-sweep energy densities at 0.5C and 3C. The normalized components are
\[
s_{\mathrm{low}} =
\frac{\mathrm{ED}(0.5\mathrm{C})}{Q_{\mathrm{low}}},
\qquad
s_{\mathrm{high}} =
\frac{\mathrm{ED}(3\mathrm{C})}{Q_{\mathrm{high}}}.
\]
The scalar score optimized by BO is
\[
\mathrm{score}
=
\frac{2s_{\mathrm{high}}s_{\mathrm{low}}}
{s_{\mathrm{high}}+s_{\mathrm{low}}}
+0.25\,\mathbb{I}
\left[
s_{\mathrm{high}}\ge 1.0
\ \mathrm{and}\
s_{\mathrm{low}}\ge 0.9
\right].
\]
Higher score is better. The harmonic-mean term rewards designs that perform well at both rates and penalizes one-sided solutions; the bonus rewards candidates that simultaneously clear the high-rate and low-rate thresholds.

\paragraph{Pool and budget.} We evaluate BO on a fixed discrete candidate pool of 200 designs sampled from a larger PyBaMM sweep. The pool is deliberately imbalanced to mimic sparse high-performing regions: 5 high-score designs have scores at least 0.95, 25 intermediate designs have scores between 0.50 and 0.90, and 170 low-score decoys have scores at most 0.20. The maximum score in the pool is 1.2123. Therefore, only 5/200 = 2.5\% of candidates exceed the 0.95 high-score threshold, and only 2/200 = 1.0\% achieve a score of at least 1.0.

Each run starts from 3 initial observations and performs up to 20 sequential one-point BO acquisitions over the remaining candidates. A run is terminated early once the global maximum in the pool is observed. For plotting only, terminated trajectories are padded with their final best-so-far value so that all runs have the same length.

\paragraph{Methods and acquisition.}
We compare Random, GP+EI, TPE, SMAC3, LLM Pointwise Neutral, and LLM Pointwise Warning. The LLM model is GPT-5.4. The LLM sees the active physical variables, bounds, the fixed value of \(R_{\mathrm{pos}}\), observed designs with score and rate-wise components, and one candidate at a time. As in the HPO experiment, LLM uncertainty is derived from token-level uncertainty, and all non-random methods use expected improvement.

\paragraph{Representative LLM prompts.}
We show concise templates rather than full run-specific candidate logs.

\paragraph{\pw{} \neutral{}.}
\begin{quote}
\small
You are optimizing a battery cathode design with two active variables:
\(L_{\mathrm{pos}}\) and \(\epsilon_{\mathrm{pos}}\). 
Maximize energy density under both low-rate (0.5C) and high-rate (3C) conditions.
Prefer high and balanced performance across both rates.\\
\\
Score:\\
\(s_{\mathrm{low}}=\mathrm{ED}(0.5\mathrm{C})/Q_{\mathrm{low}}\)\\
\(s_{\mathrm{high}}=\mathrm{ED}(3\mathrm{C})/Q_{\mathrm{high}}\)\\
\(\mathrm{score}=2s_{\mathrm{high}}s_{\mathrm{low}}/(s_{\mathrm{high}}+s_{\mathrm{low}})
+0.25\,\mathbb{I}[s_{\mathrm{high}}\ge1.0 \ \mathrm{and}\ s_{\mathrm{low}}\ge0.9]\).\\
Higher score is strictly better.\\
\\
Observed trials so far:\\
\texttt{[1]} [design 1] \(\rightarrow\) score = [value 1]
(\(s_{\mathrm{low}}=\) [low-rate component],
\(s_{\mathrm{high}}=\) [high-rate component])\\
\ldots\\
\\
Candidate to score:\\
\emph{candidate design}\\
\\
Predict the expected \texttt{score} for this candidate using the formula above.\\
Return only one numeric value.
\end{quote}

\paragraph{\pw{} Warning.}
\begin{quote}
\small
The warning condition uses the same neutral prompt with the following additional block:\\
\\
IMPORTANT PHYSICS WARNING\\
PAY CLOSE ATTENTION TO THIS. DO NOT IGNORE THE RATE-DEPENDENT PHYSICS:\\
- Low-rate energy density improves with lower inactive content, such as lower porosity.\\
- At high rate, transport pathways and porosity become more important.\\
Use this warning aggressively when judging candidates. A design that looks good at only one rate is not enough; the score rewards high, balanced performance across both low-rate and high-rate discharge.
\end{quote}

\paragraph{Outcome summary.}
Figure~\ref{fig:battery} reports best-so-far regret relative to the pool optimum. The Warning prompt gives the strongest final performance: its mean final regret is \(0.0500\), it reaches the treasure threshold \(0.95\) in all \(10/10\) seeds, and it finds the exact global optimum in \(8/10\) seeds. SMAC3 is the closest non-LLM competitor, with mean final regret \(0.0574\), treasure-threshold hit rate \(9/10\), and global-optimum hit rate \(6/10\).

The Neutral prompt also improves substantially over other non-LLM search, reaching the treasure threshold in \(9/10\) seeds and the global optimum in \(5/10\) seeds, with mean final regret \(0.1310\). The remaining baselines have larger mean final regret: \(0.2284\) for TPE, \(0.2696\) for Random, and \(0.2718\) for GP+EI. Although TPE and GP+EI occasionally find the optimum, their high variance leads to worse mean final regret. Overall, the result shows that the pointwise LLM surrogate can be competitive on this sparse fixed-pool battery-design problem, and that adding the physics warning yields the most reliable performance.

\section{Licenses and Existing Assets}

We use existing language models, software packages, datasets, and benchmark objectives only for evaluation. OpenAI models are accessed via the OpenAI API and are subject to OpenAI's Terms of Use. Qwen models are used under their respective open-source licenses (e.g., Apache License 2.0 for certain releases), and Llama models are used under the Meta Llama Community License Agreement.

We use XGBoost, Optuna, SMAC3, PyBaMM, NumPy, SciPy, scikit-learn, and Matplotlib under permissive open-source licenses (e.g., BSD, MIT, Apache 2.0). The California Housing dataset is accessed through scikit-learn as a standard public regression benchmark. The Branin objective is a standard synthetic benchmark, and the shifted Branin and battery-design candidate pool are generated as described in Appendix~\ref{app:details}.

We do not redistribute third-party model weights, datasets, or software packages. All assets are used in accordance with their respective licenses or terms of use.

\section{Key implementation constants}
\label{app:key-implementation-constants}

Table~\ref{tab:key-implementation-constants} summarizes the main implementation
constants needed to reproduce the ambiguity profiles, metric scales, GP controls, and
LLM-based acquisition scores. These constants were fixed before evaluation and were
not tuned separately for individual prompts, protocols, or models. For LLM-based BO,
token-level negative log-probability is used as a heuristic exploration feature rather
than as a calibrated objective-scale posterior standard deviation.

\begin{table}[h]
\centering
\caption{Key implementation constants.}
\label{tab:key-implementation-constants}
\small
\begin{tabular}{p{0.3\textwidth} p{0.22\textwidth} p{0.4\textwidth}}
\toprule
Component & Quantity & Value / setting \\
\midrule

Reference ambiguity profiles
& Query grid
& $[-5,5]$ with 41 query points; observed anchors handled separately \\
FRS
& Scale parameter
& $\tau_f=\operatorname{IQR}(y)^2$; aggregate protocol run used $\tau_f=0.64$ \\

Sequential GP control
& GP kernel and nugget
& Constant kernel amplitude $1.0$ times RBF length-scale $1.0$; nugget $10^{-6}$ \\

Sequential information score
& Local weighting
& Global/local weights $0.35/0.65$; local Gaussian bandwidth $1.5$ \\

Conflict construction
& Conflict strength
& $\lambda=3.0$ and $\Delta_{\min}=\max(0.8,0.35\lambda)$ \\

LLM token uncertainty
& Aggregation
& Mean negative log-probability over numeric-output tokens \\

Branin BO
& LLM acquisition score
& $\mu_t^{\mathrm{LLM}}(x)+3.0\,n_t(x)$ \\

HPO and battery BO
& LLM uncertainty feature
& Mean token NLL used as heuristic uncertainty feature, floored at $10^{-6}$ \\

\bottomrule
\end{tabular}
\end{table}

Here, $\operatorname{IQR}(y)$ denotes the interquartile range of the target values used
to set the FRS scale for the corresponding aggregate evaluation, and $n_t(x)$ denotes
the mean token negative log-probability of the numeric output.

%%%%%%%%%%%%%%%%%%%%%%%%%%%%%%%%%%%%%%%%%%%%%%%%%%%%%%%%%%%%

% \newpage
% \input{checklist.tex}

\end{document}